\documentclass{article}

\usepackage{arxiv}
\usepackage[utf8]{inputenc} % allow utf-8 input
\usepackage[T1]{fontenc}    % use 8-bit T1 fonts
\usepackage{hyperref}       % hyperlinks
\usepackage{url}            % simple URL typesetting
\usepackage{booktabs}       % professional-quality tables
\usepackage{amsfonts}       % blackboard math symbols
\usepackage{nicefrac}       % compact symbols for 1/2, etc.
\usepackage{microtype}      % microtypography
\usepackage{graphicx}
\usepackage{doi}
\usepackage{adjustbox}
\usepackage{algorithm}
\usepackage{algpseudocode}
\usepackage{amssymb}
\usepackage{natbib}
\usepackage{soul}

\title{Enhancing Interpretable Object Abstraction via Clustering-based Slot Initialization}

\author{Ning Gao$^{1, 2}$\thanks{Accepted by British Machine Vision Conference
 (BMVC), 2023} \ \   Bernard Hohmann$^{2}$ \ \  Gerhard Neumann$^{2}$\\
$^{1}$Bosch Center for Artificial Intelligence \ \  $^{2}$Autonomous Learning Robots, KIT\\
{\tt\small ning.gao@de.bosch.com} \ \ {\tt\small bernard.hohmann@student.kit.edu}\\
{\tt\small gerhard.neumann@kit.edu}
}

\hypersetup{
pdftitle={Enhancing Interpretable Object Abstraction via Clustering-based Slot Initialization},
pdfauthor={Ning Gao, Bernard Hohmann, Gerhard Neumann},
pdfkeywords={Slot-attention, clustering, conditional representation},
}

\begin{document}
\maketitle

\begin{abstract}
Object-centric representations using slots have shown the advances towards efficient, flexible and interpretable abstraction from low-level perceptual features in a compositional scene. Current approaches randomize the initial state of slots followed by an iterative refinement. As we show in this paper, the random slot initialization significantly affects the accuracy of the final slot prediction. Moreover, current approaches require a predetermined number of slots from prior knowledge of the data, which limits the applicability in the real world. In our work, we initialize the slot representations with clustering algorithms conditioned on the perceptual input features. This requires an additional layer in the architecture to initialize the slots given the identified clusters. We design permutation invariant and permutation equivariant versions of this layer to enable the exchangeable slot representations after clustering. Additionally, we employ mean-shift clustering to automatically identify the number of slots for a given scene. We evaluate our method on object discovery and novel view synthesis tasks with various datasets. The results show that our method outperforms prior works consistently, especially for complex scenes. 
\end{abstract}

% keywords can be removed

\section{Introduction}
\label{introduction}
Object-centric representations using slots have shown good performance in object detection \citep{Li2021SCOUTERSA, SlotAttention}, segmentation \citep{SIMONe, IODINE} and tracking \citep{Wu2021APEXUO, kipf2022conditional, MULMON} tasks. Slots are a set of latent variables. The common approach is to frame disentangled and structured slot representations of the compositional scene with some iterative refinement mechanisms in a self-supervised manner, e.g., using softmax-based attention \citep{SlotAttention} or variational inference \citep{IODINE}. The idea is to improve the sample efficiency and generalization of capturing the structured environment to unseen compositions or objects. However, most slot-based approaches have difficulties in representing complex scenes. Moreover, the number of slots needs to be specified beforehand on each dataset, which limits the generalization across datasets. In addition, a random slot initialization from a common distribution is widely used in prior works, which lacks consideration between the slots and the perceptual input. Consequently, the quality of the following iterative slot refinement is also affected by the sub-optimal initialization.

\begin{figure*}
\centering
\includegraphics[width=0.9\linewidth]{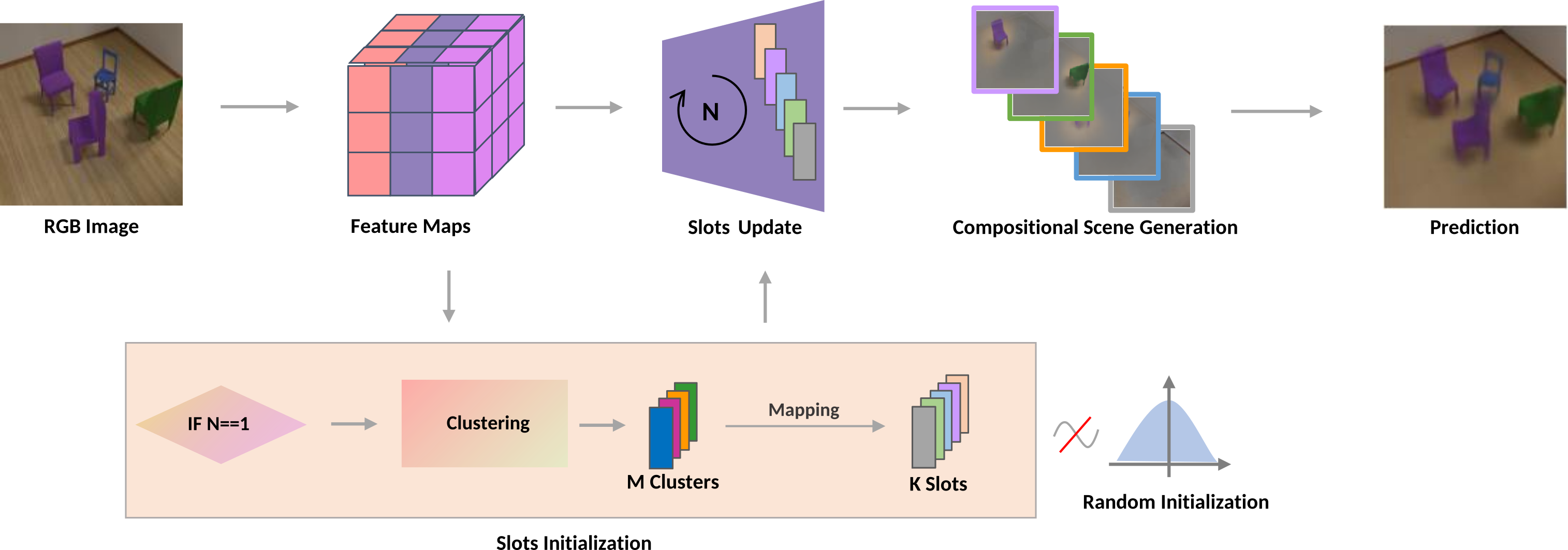}
\caption{The network architecture. Instead randomizing slot initialization from a common distribution widely used in prior work, we initialize slot representations conditioned on the input features. A clustering algorithm and a mapping layer are adopted.}
\label{fig_architecture}
\end{figure*}

To overcome these challenges, instead of random sampling, it is intuitive to sample the initial slot representations conditioned on the perceptual input (see Figure \ref{fig_architecture}). Hence, we employ the k-means clustering algorithm on the convolutional features of the input image. A set of cluster centers are specified based on the features. Afterwards, a set of slots are initialized given the cluster centers as input. Since the order of cluster centers changes randomly, we extend this idea with a permutation-invariant mechanism, where the initial slot representations remain invariant w.r.t. the order of clusters. To further evaluate the effect of permutation symmetry for slot representations, we employ another permutation equivariant model with mean-shift clustering algorithm, where the slot representations change accordingly with respect to the permutation of the clusters. Mean-shift identifies the number of clusters automatically based on each perceptual input, followed by an injective mapping where each slot is considered as an output of each cluster individually. Thus, it does not require a fixed number of slots based on the whole dataset as prior works. 

Our proposed method can be easily placed on top of existing slot-based approaches and trained in an end-to-end manner. In this work, we consider object discovery and novel view synthesis as downstream tasks. To evaluate the improvement and versatility of our method, we choose Slot Attention \citep{SlotAttention} and IODINE \citep{IODINE} as baselines for object discovery task, and uORF \citep{uORF} for novel view synthesis. The experiments are conducted on various datasets.

Our main contributions are as follows: i) We propose the conditional slot initialization using clustering algorithms instead of random initialization. ii) We analyze the effect of permutation symmetry including invariance and equivariance on the object-centric slot representations. iii) We apply mean-shift clustering on the perceptual features which allows to generate flexible number of slots. iv) We demonstrate that, our proposed idea achieves significant improvement over all baselines, while the permutation equivariant mean-shift model presents notable advances especially for complex scenes.

\section{Guiding Slot Initialization using Clustering}
\label{method}
In this section, we will introduce i) the conditional slot initialization with k-means clustering (KM) in Section \ref{method_CSI}, ii) the permutation invariant version named \textit{Pseudoweights} (PW) in Section \ref{method_PI_CSI}, iii) and the permutation equivariant version with variable slot generation using the mean-shift clustering (MS) in Section \ref{method_PE_CSI}. More details about implementations and architectures are shown in appendix A.

\subsection{Image-Dependent Slot Initialization}
\label{method_CSI}
Most slot-based methods typically sample from a standard Gaussian as the random initialization for the slot latent variables (see Figure \ref{fig_architecture}). Although the slots are updated by the refinement mechanism incorporating the features from the perceptual input, it is inefficient to start from a random initialization and also limits the final accuracy. Since the perceptual input includes a strong inductive bias about the represented scene, it is straightforward to incorporate the perceptual input directly from the beginning. We first implement a non-permutation symmetric model using k-means clustering. K-means is applied on the pixel-wise convolutional perceptual feature $\textbf{x} \in \mathbb{R}^{N \times D}$ to get the feature-based cluster centers: $\textbf{c}=\textrm{K-means}(\textbf{x}) \in \mathbb{R}^{M \times D}$ where $N$ is number of pixels from the feature input, $M$ is the number of clusters and $D$ is the feature dimension. Afterwards, the cluster centers are flattened and mapped to the $K$ slots using multi-layer perceptrons (MLPs): $\textbf{z}_{slots}=\textrm{MLP}(\textbf{c}.\textrm{flat}()).\textrm{reshape}(K, D)$. Therefore, the number of slots is fixed beforehand like in prior works, as well as the amount of cluster centers. 
% Some cluster centers can vanish during the iterative updates of k-means. For this reason, we resample new clusters during iteration, such that the output of k-means is consistently in $\mathbb{R}^{M \times D}$ and compatible with the MLPs mapping.

\subsection{Permutation-Invariant Slot Initialization}
\label{method_PI_CSI}
A good slot representation respects the permutation symmetry \citep{SlotAttention}. In our case, the order of the predicted slots should either remain the same (permutation invariance) w.r.t. the permutation of the cluster centers or change correspondingly in the same order as the cluster centers (permutation equivariance). Such symmetric behavior enables good generalization of slot representations to unseen world and objects. However, a simple mapping between $M$ cluster centers and $K$ slots as shown in Section \ref{method_CSI} breaks the permutation symmetry and cannot generalize to more slots during evaluation for the scenes with more objects. To address this issue, we propose a permutation invariant model named \textit{Pseudoweights}. To identify different slots, we use a sine-cosine positional encoding $\textbf{p}_k$ for the $k$-th slot as follows:
\begin{equation}
    \textbf{p}_k = \left(\sin\big(\frac{\pi}{D'}k\big), \cos\big(\frac{\pi}{D'}k\big), \sin\big(\frac{2\pi}{D'}k\big), \cos\big(\frac{2\pi}{D'}k\big),...,\sin\big({\pi}k), \cos({\pi}k\big)\right), \ k=1,...,K,
\end{equation}
where $D' = \frac{D}{2}$ and $D$ denotes the embedding length.
Afterwards, the cluster centers are broadcasted along the slot dimension $\textbf{c} \in \mathbb{R}^{K \times M \times D}$ and are concatenated with the broadcast of the positional encoding $\textbf{p} \in \mathbb{R}^{K \times M \times D}$ to predict the  weights $\textbf{w}=\textrm{MLPs}([\textbf{c}, \textbf{p}]) \in \mathbb{R}^{K \times M \times D}$, which allocate the importance of the cluster centers to the different slots. We use a soft-max layer such that the weights allocated for each slot are normalized as follows:
% $$\sum_{i=1}^{M} w_{im} = 1, \,  w_{im} \in [0, 1] \subset \mathbb{R}.$$
\begin{equation}
    \sum_{m=1}^{M} w_{k,m,d} = 1, \,   w_{k,m,d} \in [0, 1], \, k=1,...,K,\, m=1,...,M,\ d=1,...,D.
\end{equation}
The slots are then initialized as the weighted sum over the cluster centers by $\textbf{w}$:
% $$ z_{m} = \sum_{m=1}^{M} w_{im} \cdot c_{m};\, i=1,...,D.$$
% $$ z_{k} = \sum_{m=1}^{M} w_{im} \cdot c_{m};\, i=1,...,D;\, k=1,...,K.$$
\begin{equation}
    \textbf{z}_{k} = \sum_{m=1}^{M} \textbf{w}_{k,m} \cdot \textbf{c}_{k,m}.
\end{equation}
 Thus, the \textit{Pseudoweights} mapping applies a permutation invariant assignment of cluster centers into the slots. Moreover, since the slots are identified by the positional encoding, it enables generalization on increasing objects during test by changing the defined number of slots $K$ without increasing the model parameters. A detailed visualization of the architecture is depicted in appendix A. 
% The network decides conditioned only on one particular cluster center and one $\textbf{p}_k$, if a cluster center is a member of a slot. This process is repeated for all $M \times K$ possible combinations of (cluster center, position code)-combinations, resulting in a weight tensor, that has high weights, if a cluster center is a member of a slot, otherwise low weights. The resulting slot initializations $\textbf{z}$ are calculated with a weighted sum of all members $\textbf{c}_{k}$ of a slot with weights $w_{ik}$: 
% \begin{equation}
% \begin{gathered}
%     z_{m} = \sum_{m=1}^{M} w_{im} \cdot c_{m};\, i=1,...,D\\
%     1 = \sum_{i=1}^{D} w_{im};\,  w_{im} \in [0, 1] \subset \mathbb{R}
% \end{gathered}
% \end{equation}
% With the linear mapping we also make sure that the slot initializations $\textbf{z}$ are in the same space as the pixels in the perceptual input $\textbf{x}$. In contrast to that, it can not be easily proven for non linear mappings like the method in section \ref{method_CSI}, that $\textbf{z}$ are initialized in the same space as $\textbf{x}$. Thus making it easier for the mapping algorithm to initialize slots closer to the perceptual input.
\subsection{Automatic Tuning of the Number of Slots using Mean-Shift}
\label{method_PE_CSI}
Both models introduced in Section \ref{method_CSI} and Section \ref{method_PI_CSI} still require a fixed number of slots beforehand. Therefore, it is essential to apply an unsupervised clustering mechanism to determine the number of slots conditioned on the input features while keeping the permutation symmetry. Consequently, we perform the mean-shift clustering algorithm \citep{meanshift} over the feature space to determine the cluster centers. Mean-shift is an iterative procedure to approximate different modes of a distribution using kernel density estimation. Each mode is represented as a cluster which does not need to be determined beforehand. In our model, we use a Gaussian kernel $k(x, y) = \exp(-\frac{1}{\sigma^2}||x-y||^2)$ for the density estimation. $\sigma$ is a hyperparameter which affects the granularity of the modes. A shared mapping layer is utilized to initialize the slots based on each cluster respectively $\textbf{z}_i={\textrm{MLP}}_{\textrm{shared}}(\textbf{c}_i)$ where $i \in \{1, ..., K\}$. Thus, it holds the permutation equivariance but requires to have the same number of slots as the number of the predicted cluster centers $K=M$. Since the Gaussian kernel is predefined by a hyperparameter, an expressive learned convolutional feature space is crucial to output distinctive modes. 
% This is achieved by a fully differentiable pipeline including the fully differentiable implementations of the mean-shift algorithm.
\section{Related Work}
\label{relatedwork}
\noindent\textbf{Object-centric slot representations.} Slot representations have been widely used in static scenes \citep{SlotAttention, IODINE, DETR, MONet, GENESIS} and videos \citep{Spatial_MULMON, Yang_2021_ICCV, kipf2022conditional, OP3, ViMONET}. Each slot represents a corresponding object in the scene. This can be achieved either by accumulating the evidence over time to maintain the consistent object slot \citep{ViMONET} if a variational auto-encoder \citep{AutoEncoder} is employed, or using softmax-based attention mechanism \citep{SlotAttention, Bao_2022_CVPR}. However, all of these approaches require a fixed set of slot variables. The set size needs to be strictly equal or larger than the number of objects in the scene, which limits the generalization on real-world applications since the number of objects is changing dynamically over time and cannot be determined in advance.

\noindent\textbf{Scene decomposition.} Most works formulate scene decomposition as compositional generative model \citep{IODINE, NIPS2016_52947e0a, vonKgelgen2020TowardsCG} or a mixture of components \citep{SlotAttention, MONet, GENESIS}. Recently, some works \citep{Stelzner2021Decomposing, uORF, wang2022dmnerf} extend 2D scene decomposition to 3D with the advances of Neural Radiance Field (NeRF) \citep{NeRF}. \cite{Chen2020LearningTI} and \cite{MULMON} infer 3D scenes from multiple reference images and textureless background. In contrast, uORF \citep{uORF} infer from a single image and test on complex objects with diverse textured background.

\noindent\textbf{Object discovery.} Object discovery requires to differentiate the objects and background in an unsupervised way. These methods typically model objects as a set of latent embeddings \citep{DETR} and adopt topic modelling \citep{Russell2006UsingMS}, group image patches \citep{Tuytelaars2009UnsupervisedOD, Grauman2006UnsupervisedLO} or clustering-based deep learning algorithms \citep{Li2019GroupWiseDO, Saliency}. Some methods \citep{Pyramid, NEURIPS2021_8bf1211f} also apply saliency detection and region proposals on the entire image to group and localize the objects. 

\noindent\textbf{Novel view synthesis.} Novel view synthesis aims to generate novel views of the given scene from a single \citep{IODINE, GQN, uORF} or multiple \citep{MULMON, NeRF} source views. \cite{Liu2021NovelVS} employ a token-transformation module to synthesize the novel views from a single image without requiring the pose information. \cite{Chen2021STRGQNSR} extend GQN \citep{GQN} with a Spatial Transformation Routing (STR) mechanism without requiring explicit camera intrinsic information. \cite{VMV} enable the real-time novel view inference with the advantage of volume rendering. Recently, \cite{FWD} replace the expensive computation of volumetric sampling in NeRF-like methods by pixel-wise depth prediction and a differentiable point cloud renderer.

\noindent\textbf{Deep clustering.} 
% Clustering is central to many data-driven research and unsupervised learning. In particular, 
Clustering helps analyze unstructured and high-dimensional data into meaningful and low-dimensional representations, which has been improved with deep learning techniques in recent years \citep{DEC_clustering}. \cite{AE_clustering} propose an iterative optimization of learning low-dimensional representations from an auto-encoder by minimizing the Kullback-Leibler divergence between the pixel-wise features to each cluster center. \cite{Entropy_clustering} extend it with a classifier on top which predicts the probability over the $k$ classes where $k$ is the number of cluster centers. \cite{deep-kmeans} employ the objective of k-means as clustering loss in the feature space while \cite{soft-deep-kmeans} relax the cluster assignment problem by using a soft-assignment which can fully benefit from the efficiency of stochastic gradient decent (SGD). \cite{Genevay2019DifferentiableDC} propose a fully differentiable version with the cluster parameters while \cite{Cai_2022_CVPR} reduce the computational time by introducing a subspace-based clustering and improve the scalability of deep clustering.

\section{Experiments}
\label{experiments}
To evaluate our method, we choose two object-centric tasks: object discovery in Section \ref{sec-object-discovery} and novel view synthesis in Section \ref{sec-novel-view-synthesis}. We employ our idea on top of three state-of-the-art methods: Slot Attention \citep{SlotAttention}, IODINE \citep{IODINE} and uORF \citep{uORF}. We show more details about implementations in appendix A and qualitative results in appendix B and C.

\noindent\textbf{Baselines.} In the object discovery task, we use Slot Attention and IODINE as baselines and build our method on top of them. Both baselines use slot representations but with different procedures to refine the slots: Slot Attention uses simple but effective softmax-based attention mechanism while IODINE considers slots as probabilistic latent variables and employs variational inference to accumulate the evidence during iterations. For the novel view synthesis task, we choose uORF as baseline which uses softmax-based attention module to update slots and generate slot-based compositional scenes with Neural Radiance Field (NeRF). Note that all these models use random slot initialization. In addition, we also design two ablated models where the slot initialization is conditioned on the input features. First, we employ the k-means initialization directly as slot representations without any mapping layers in between (\textit{direct} model). Second, we design a simple and permutation equivariant model using shared MLPs to map the k-means cluster centers of the input features to the slots (\textit{shared MLPs} model). 

\noindent\textbf{Datasets.} We use three datasets for the object discovery task:  Multi-dSprites (MDS), CLEVR and Chairs datasets. Each dataset contains multiple objects in the scene. Similar as Slot Attention, we extract the CLEVR dataset to have maximum 4, 6, and 10 objects respectively and denote them as CLEVR4, CLEVR6 and CLEVR10. The Chairs dataset originates from uORF\citep{uORF}, which includes 4 chairs in each scene. The dataset includes 1200 different shapes of chairs sampled from ShapeNet \citep{ShapeNet} and 50 different floor textures as background. To train the Slot Atttention related models, we use 5k images for CLEVR4 and 10k for MDS, CLEVR6 and Chairs. To train the IODINE related models, we use the same datasets except 13.9k images for MDS. Each dataset contains another 500 images for evaluation. For the novel view synthesis task: We only use the Chairs dataset but it includes 5k scenes for training and 500 scenes for testing, where each scene includes 4 images with different camera viewpoints. Therefore, there are in total 20k images for training and 2k images for testing.

\noindent\textbf{Metrics.} As prior works \citep{IODINE, SlotAttention, MONet}, for the object discovery task, we adapt the Adjusted Rand Index (ARI) score to be evaluated only on the pixels of the foreground objects and evaluate the predicted segmentation with the groundtruth mask. For the novel view synthesis, we follow uORF and adopt ARI on the fully reconstructed image, the foreground regions (Fg-ARI) and the synthesized novel view images (NV-ARI). Furthermore, we use LPIPS \citep{LPIPS}, SSIM \citep{SSIM} and PSNR \citep{PSNR} as perceptual metrics for both tasks.

\subsection{Object Discovery}
\label{sec-object-discovery}

\noindent\textbf{Training.} We follow the same training setup of Slot Attention and IODINE. We use Adam optimizer \citep{Kingma2015AdamAM} with a learning rate of $4\times10^{-4}$ for Slot Attention based models and $3\times10^{-4}$ for IODINE related models. We train the Slot Attention related models with 2 NVIDIA Tesla V100-32GB GPUs and a batch size of 32 on each GPU. For IODINE related models, we use 4 GPUs since IODINE requires more computation and gpu memory. We train each model for 1000 epochs with a warm-up training strategy \citep{WarmUp} and an exponential learning rate decay. We use $K=5$ for MDS, CLEVR4 and Chairs datasets since there are maximum 4 objects in each scene, and $K=7$ for CLEVR6. The cluster number is set to $M=2K$ except for the \textit{mean-shift}, \textit{direct} and \textit{shared MLPs} versions which require $M=K$.

\begin{table}[t]
\caption{Quantitative results on the object discovery task.}
\label{tab-object-discovery}
\begin{center}
% \begin{tabular}{@{}l|c*{5}c|c*{5}c|c*{5}c|c*{5}c|c*{5}c@{}}\toprule \\
\begin{adjustbox}{width=1\textwidth}
{\begin{tabular}{@{}l|ccccc|ccccc|ccccc@{}}\toprule
 & \multicolumn{5}{c|}{\textbf{MDS}}  & \multicolumn{5}{c|} {\textbf{CLEVR6}} & \multicolumn{5}{c}{\textbf{Chairs}}\\
% \\\midrule
\textbf{Model}  & ARI $\uparrow$ & LPIPS$_A$ $\downarrow$ & LPIPS$_V$ $\downarrow$ & PSNR $\uparrow$ & SSIM $\uparrow$ & ARI $\uparrow$ & LPIPS$_A$ $\downarrow$ & LPIPS$_V$ $\downarrow$ & PSNR $\uparrow$ & SSIM $\uparrow$ & ARI $\uparrow$ & LPIPS$_A$ $\downarrow$ & LPIPS$_V$ $\downarrow$ & PSNR $\uparrow$ & SSIM $\uparrow$
\\\midrule
SL & 0.9671 & 0.0693 & 0.1351 & 27.43 & 0.9237 %MDS
   & 0.9815 & 0.0748 & 0.1486 & 32.11 & 0.8908 % CLEVR6
   & 0.9982 & 0.3144 & 0.4362 & 24.49 & 0.6035 % Chairs_D
   \\
SL + kmeans (direct) & 0.9223 & 0.1074 & 0.1606 & 26.13 & 0.9095 % MDS
                     & 0.9963 & 0.0381 & 0.1097 & 34.22 & 0.9161 % CLEVR6
                     & 0.9963 & 0.2971 & 0.4273 & 24.17 & 0.6024 % Chairs_D
\\
SL + kmeans & 0.9837 & 0.0519 & 0.1149 & 28.88 & 0.9417 % MDS
            & 0.9970 & 0.0313 & 0.1032 & 34.98 & 0.9255 % CLEVR6
            & 0.6271 & 0.2948 & 0.4274 & 24.31 & 0.6034 % Chairs_D
            \\
SL + kmeans (shared MLPs) & 0.9043 & 0.1174 & 0.1672 & 25.84 & 0.9019 % MDS
                         & 0.9989 & 0.0320 & 0.1041 & 34.82 & 0.9255 % CLEVR6
                         & 0.9974 & 0.3173 & 0.4297 & 25.01 & 0.6199 % Chairs_D
                         \\
SL + PW & 0.9605 & 0.0834 & 0.1526 & 26.25 & 0.9104 %MDS
        & 0.9937 & 0.0371 & 0.1056 & 34.04 & 0.9251 % CLEVR6
        & 0.9523 & 0.3052  & 0.4363 & 24.82 & 0.6104 % Chairs_D
        \\
SL + MS (direct) & 0.9893 & 0.0448 & 0.1059 & 31.39 & 0.9559 % MDS
                 & 0.6114 & 0.1098 & 0.1957 & 29.43 & 0.8555 % CLEVR6
                 & 0.9999 & 0.2757 & 0.3997 & 26.02 & 0.6341\\
SL + MS & \textbf{0.9944} & \textbf{0.0393} & \textbf{0.0919} & \textbf{32.17} & \textbf{0.9613} %MDS
        & \textbf{1.0000} & \textbf{0.0306} & \textbf{0.1022} & \textbf{35.32} & \textbf{0.9301} % CLEVR6
        & \textbf{1.0000} & \textbf{0.2693} & \textbf{0.3774} & \textbf{26.03} & \textbf{0.6444} % Chairs_D
        \\
ID & 0.9362 & 0.0504 & 0.0888 & 30.91 & 0.9591 %MDS
   & 0.8990 & 0.0224 & 0.0500 & 37.5 & 0.9661 % CLEVR6
   & 0.2185 & 0.2757 & 0.3843 & 24.27 & 0.6299 % Chairs_D
   \\
ID + kmeans (direct) & 0.9910 & 0.0193 & 0.0492 & 36.03 & 0.9833 %MDS
                     & 0.8791 & 0.0254 & 0.0559 & 36.86 & 0.9619 % CLEVR6
                     & 0.6881 & 0.2666 & 0.3842 & 24.25 & 0.6322 % Chairs
                     \\
ID + kmeans & 0.9962 & 0.0166 & 0.0415 & 37.06 & 0.9861 %MDS
            & 0.8325 & 0.0198 & 0.0479 & 37.725 & 0.9667  % CLEVR6
            & 0.7281 & 0.2559 & 0.3744 & 24.31 & 0.6314 % Chairs_D
            \\
ID + PW & 0.9930 & 0.0207 & 0.0440 & 36.42 & 0.9834 %MDS
        & 0.9818 & 0.0190 & 0.0483 & 37.725 & 0.9667 % CLEVR6
        & 0.8792 & 0.2192 & 0.3712 & 29.025 & 0.6362 \\
ID + MS & \textbf{0.9970} & \textbf{0.0143} & \textbf{0.0401} & \textbf{38.12} & \textbf{0.9921}     
        & \textbf{0.9909} & \textbf{0.0141} & \textbf{0.0361} & \textbf{38.90}  & \textbf{0.9753} % CLEVR6
        & \textbf{0.9991} & \textbf{0.1645} & \textbf{0.3219} & \textbf{31.07} & \textbf{0.6995}
        \\

\bottomrule
\end{tabular}}
\end{adjustbox}
\end{center}
\end{table}

\begin{figure*}
\centering
\includegraphics[width=0.9\linewidth]{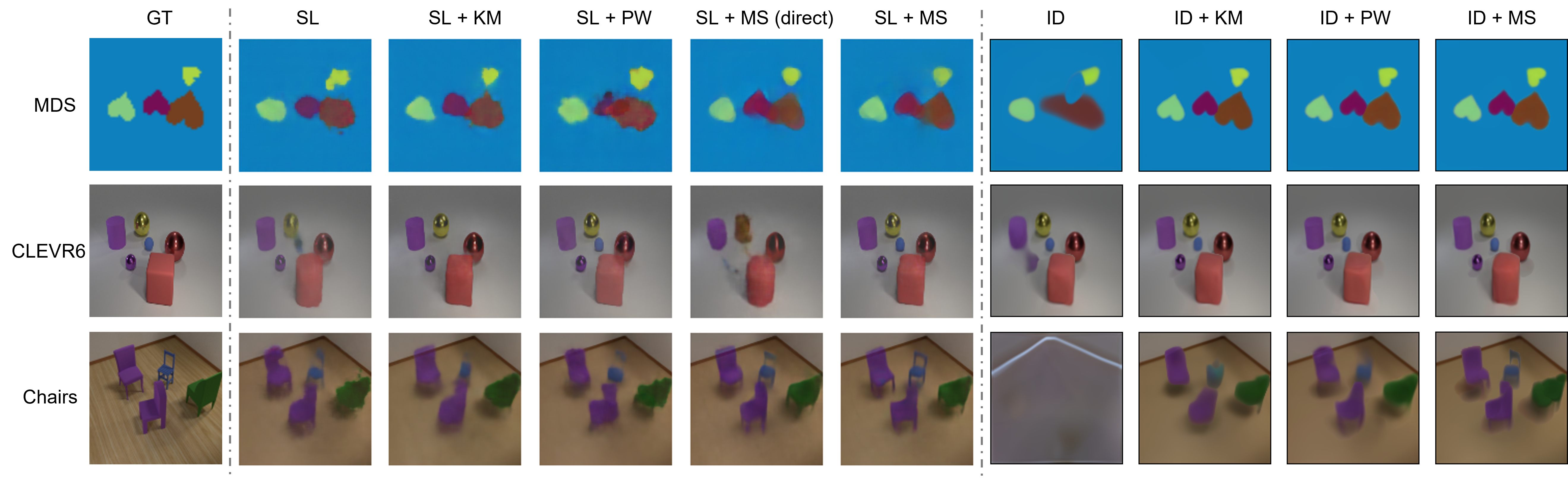}
\caption{Qualitative results on the object discovery task. Notably, the \textit{mean-shift} (MS) versions can recover detailed appearance over all datasets with even better quality than original input for IODINE-based models in MDS dataset.}
\label{fig_results_sl}
\end{figure*}

\begin{figure}[htb!]
\centering
\includegraphics[width=0.9\linewidth]{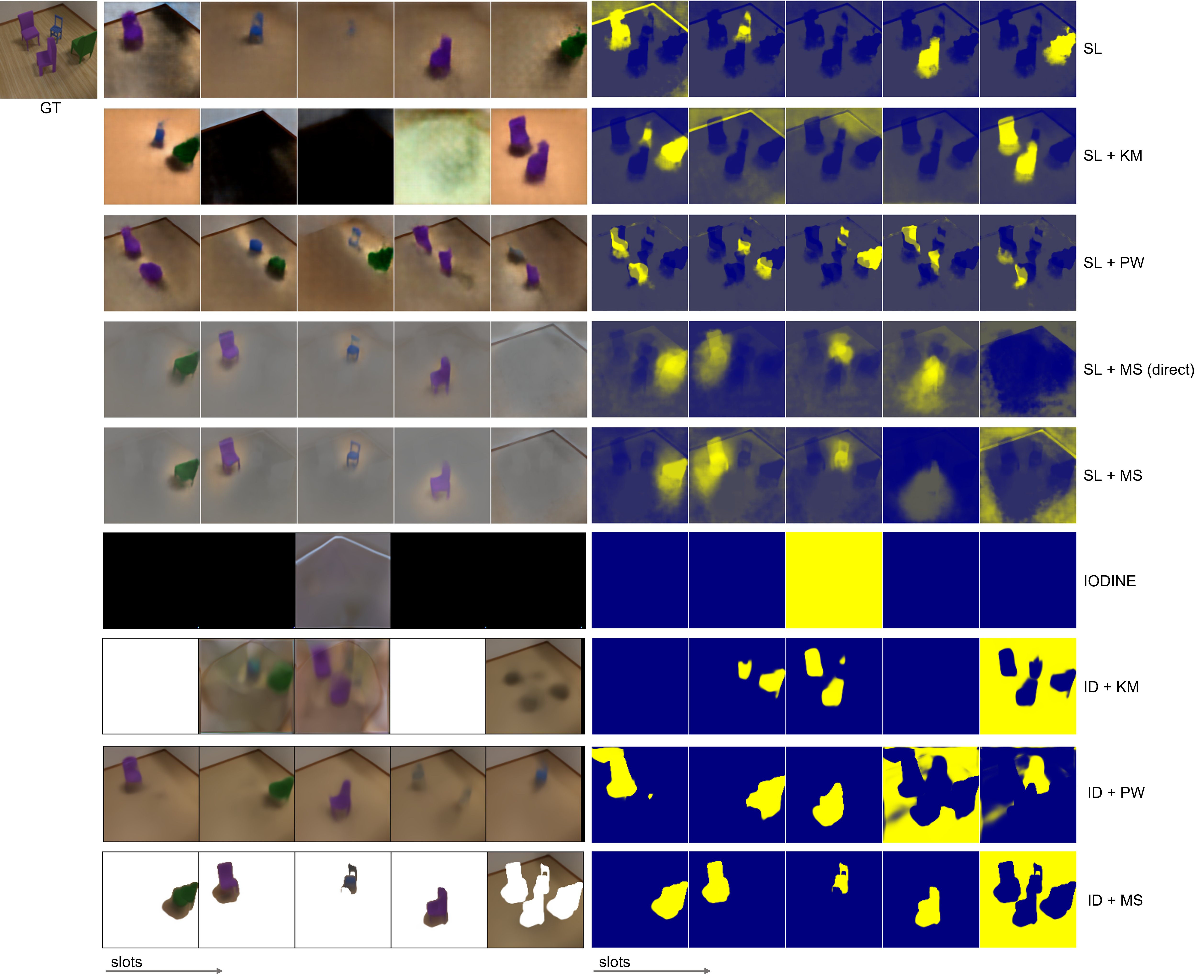}
\caption{Qualitative results of slot-wise reconstructed scenes (left) and masks (right). \textit{Mean-shift} models disentangle the objects better than others and recover more details.}
\label{fig_sl_recons_slots}
\end{figure}

\noindent\textbf{Results.} Quantitative results are shown in Table \ref{tab-object-discovery} and qualitative results in Figure \ref{fig_results_sl}. In general, learning inductive slot initialization from input features improve the performance on both baselines, where \textit{mean-shift} models achieve the best performance consistently over all datasets. \textbf{Well-recovered details:} Surprisingly, all our IODINE-based variants achieve higher resolution even than the groundtruth image for MDS dataset, while the original IODINE is struggled with the data prior and cannot reconstruct the shape of objects. Furthermore, in Figure \ref{fig_results_sl}, we observe that only the \textit{mean-shift} models can capture the details of objects for Slot Attention based models. For example, it captures the ``heart'' objects in MDS while others struggle with the data prior. In particular, our models (especially for \textit{mean-shift models}) can reconstruct the appearance in very good details, e.g., the small blue sphere in CLEVR6 and the legs and rims of various chairs in Chairs dataset. 

\textbf{Slots disentanglement:} We also visualize the slot-wise reconstructed scenes and masks in Figure \ref{fig_sl_recons_slots}. From the masks, we observe that only the \textit{mean-shift} models can fully disentangle the objects and background where the highlighted area indicates large attention. In contrast, original Slot Attention mixes the background and a chair in slot \#1 while IODINE cannot even work with textured background. \textit{Pseudoweights} and \textit{k-means} models also entangle the chairs into one slot even though the overall reconstructed performance is still better than the baselines (Table \ref{tab-object-discovery} and Figure \ref{fig_results_sl}). The slot-wise reconstructed scenes also reveal our conclusion that \textit{mean-shift} models contain more appearance details with fully disentangled slots. 

\textbf{Mapping between clusters and slots:} Furthermore, our ablation studies demonstrate that the \textit{k-means} models using non-linear mapping layers between the clusters and slots gain additional benefits compared to the \textit{direct} models (in Table \ref{tab-object-discovery}). Additionally, the permutation equivariant model (\textit{shared MLPs}) performs better than the non-permutation symmetric model (\textit{k-means}) on CLEVR6 and Chairs datasets, indicating the benefits of permutation symmetry on complex scenes, though it is not as good as the \textit{mean-shift} models especially on MDS dataset.
\begin{figure}[t]
\centering
\includegraphics[width=0.9\linewidth]{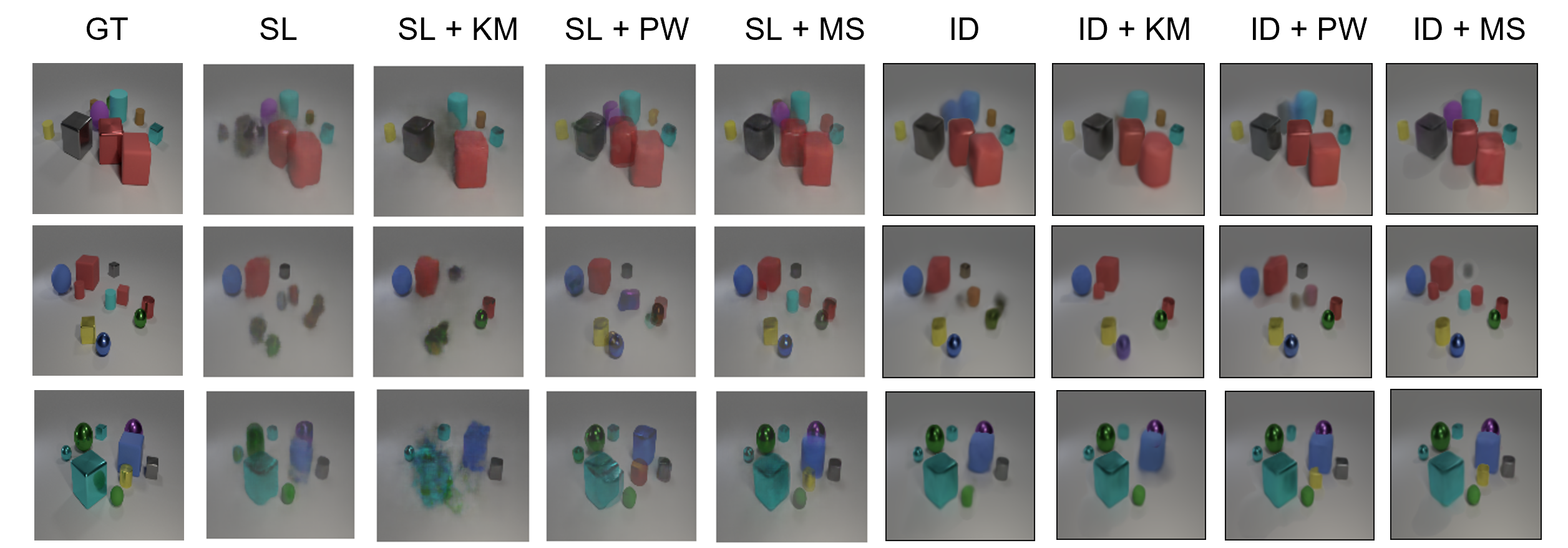}
\caption{Qualitative results on increasing objects. The models are trained on CLEVR6 but evaluated on CLEVR10 with larger number of objects.}
\label{fig_sl_generalization_clvr10}
\end{figure}
\begin{table}[t]
\caption{Evaluation with different number of iterations (5 iterations are used for training). In particular, our models achieve significant improvement already at the first iteration.}
\label{tab-different-iterations}
\begin{center}
% \begin{tabular}{@{}l|c*{5}c|c*{5}c|c*{5}c|c*{5}c|c*{5}c@{}}\toprule \\
\begin{adjustbox}{width=1\textwidth}
{\begin{tabular}{@{}l|cccc|cccc|cccc@{}}\toprule
 & \multicolumn{4}{c|}{\textbf{Iter 1}}  & \multicolumn{4}{c|} {\textbf{Iter 3}} & \multicolumn{4}{c}{\textbf{Iter 7}}\\
% \\\midrule
\textbf{Model}   & LPIPS$_A$ $\downarrow$ & LPIPS$_V$ $\downarrow$ & PSNR $\uparrow$ & SSIM $\uparrow$ & LPIPS$_A$ $\downarrow$ & LPIPS$_V$ $\downarrow$ & PSNR $\uparrow$ & SSIM $\uparrow$ & LPIPS$_A$ $\downarrow$ & LPIPS$_V$ $\downarrow$ & PSNR $\uparrow$ & SSIM $\uparrow$
\\\midrule
ID          & 0.4415 & 0.6071 & 12.72 & 0.3820 % Iter1
            & 0.4477 & 0.5804 & 16.33 & 0.4908 % Iter3
            & 0.4363 & 0.5646 & 19.53 & 0.5001 % Iter7
   \\
ID + kmeans & 0.2108 & 0.3768 & 27.05 & 0.6202 %Iter1
            & 0.1956 & 0.3607 & 28.75 & 0.6533  % Iter3
            & 0.1884 & 0.3545 & 29.33 & 0.6656 % Iter7
            \\
ID + PW     & 0.2269 & 0.3734 & 27.57 & 0.6297 % Iter1
            & 0.1973 & 0.3531 & 29.33 & 0.6642 % Iter3
            & 0.1885 & 0.3461 & 29.92 & 0.6768 % Iter7
            \\
ID + MS     & \textbf{0.1798} & \textbf{0.3545} & \textbf{28.39} & \textbf{0.6467} % Iter1
            & \textbf{0.1602} & \textbf{0.3343} & \textbf{30.16} & \textbf{0.6828} % Iter3
            & \textbf{0.1528} & \textbf{0.3273} & \textbf{30.68} & \textbf{0.6951} % Iter7
            \\
\bottomrule
\end{tabular}}
\end{adjustbox}
\end{center}
\end{table}

\textbf{Generalization on increasing objects:} In addition, we evaluate the generalization on more objects and slots (CLEVR10) while the models are trained on CLEVR6. The qualitative results are shown in Figure \ref{fig_sl_generalization_clvr10}. We observe that the original baselines struggle with closed or overlapped objects by missing, mixing or predicting wrong color of objects, while our models (especially the \textit{mean-shift} models) can detect the overlapped objects perfectly without missing any object even for extremely difficult scenes. For example, i) the \textit{mean-shift} Slot Attention model (ID + MS) can recognize all the objects in the first example with right colors and shapes, ii) in the second example, both \textit{mean-shift} models (SL + MS and ID + MS) and \textit{k-means} IODINE (ID + KM) can detect the red small cylinder in front of the red cube, though the objects are overlapped and with the same color, and iii) both \textit{mean-shift} models and \textit{Pseudoweights} IODINE (ID + PW) can reconstruct the yellow cylinder in the third example. We believe the benefits come from the inductive slot initialization conditioning on the perceptual input features, which gives expressive slot representations used in the following slot refinement. Note that \textit{k-means} models can merely detect 6 objects from the scene since the slot number is by design not scalable. 

\textbf{Generalization on increasing iterations:} Furthermore, Table \ref{tab-different-iterations} shows the evaluation with increasing number of iterations up to 7 while the models are trained with 5 iterations. All models are capable of generalizing on more iterations with performance gains. In particular, using inductive slot initialization enables notable improvement at the first iteration, which indicates the efficiency of the learned inductive slot initialization. 

\begin{figure}[t]
\centering
\includegraphics[width=0.9\linewidth]{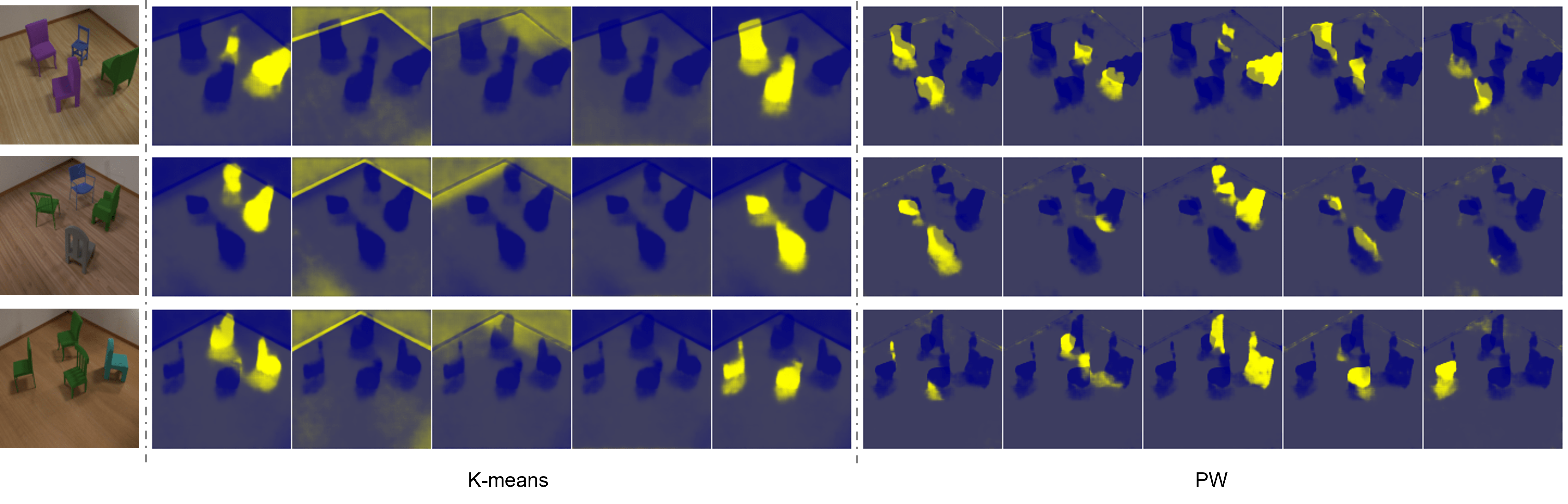}
\caption{Failure cases on Chairs dataset where \textit{k-means} and \textit{Psuedoweights} (PW) cannot disentangle the objects and use each individual slot for specific areas.}
\label{fig_sl_failuer}
\end{figure}

\textbf{Failure cases:} We further investigate the cases when \textit{k-means} and \textit{Pseudoweights} are failed to disentangle objects in Chairs dataset. Examples are shown in Figure \ref{fig_sl_failuer}. Interestingly, we find they learned structured slot representations not always based on the objects. The slot representations of \textit{k-means} model are not generalize due to the non-permutation symmetry. Thus, it always uses the same slot to represent specific area, e.g., the first slot to represent the objects in the top right area, the second and third slots for walls.  % Meanwhile, \textit{Psuedoweights} shows similar behavior by using the slots to represent specific areas individually. 
% The reasons is as follows: K-means flattens the cluster centers into a large vector and feeds it into MPLs to get the slot initialization which do not hold the permutation symmetry. Consequently, changing the cluster orders by moving the objects in the scene can drastically effect and create fully different slot initialization even though the objects and background remain the same. Thus, it limits the model to learn generalized slot representations. 
On the other hand, \textit{Pseudoweights} outputs the same slot representations while changing the object positions due to the permutation invariance. As a result, it neglects the object-centric spatial features in the scene. Thus, the model tends to reconstruct the scenes by assigning fixed spatial area to each individual slot. Such undesirable behaviors occur especially on Chairs dataset where each scene includes 4 images with changing viewpoints. In contrast, a good permutation equivariant model such as \textit{mean-shift} can alleviate this issue and decouple the objects (as shown in Figure \ref{fig_sl_recons_slots}).
\subsection{Novel View Synthesis}
\label{sec-novel-view-synthesis}
\noindent\textbf{Setup.} The Chairs dataset contains 4 images from different viewpoints of each scene. During training, we randomly pick one image from each scene as input and reconstruct the images for the other 3 viewpoints. We use the same training loss functions and strategies as uORF \citep{uORF}. uORF is a memory-extensive model which only works with a batch size of 1 on NVIDIA Tesla V100-32GB. Meanwhile, mean-shift also consumes large memory for the intermediate tensors due to its iterative optimizations. Therefore, we cannot build a mean-shift algorithm on top of uORF with our available hardware. We consider this as a limitation of our \textit{mean-shift} model.
% \begin{figure}[hbt!]
% \centering
% \includegraphics[width=0.9\linewidth]{figs/sl/uorf_results.png}
% \caption{Qualitative results on novel view synthesis. Our models can represent the chairs with more details than the original uORF.}
% \label{fig_uorf_results}
% \end{figure}
\begin{table}[htb!]
\caption{Results of novel view synthesis on Chairs-diverse.}
\label{tab-novel-view-synthesis}
\begin{center}
% \begin{tabular}{@{}l|c*{5}c|c*{5}c|c*{5}c|c*{5}c|c*{5}c@{}}\toprule \\
\begin{adjustbox}{width=0.7\textwidth}
{\begin{tabular}{@{}l|cccccc@{}}\toprule
Model & ARI $\uparrow$ & Fg-ARI $\uparrow$ & NV-ARI $\uparrow$ & LPIPS $\downarrow$ & SSIM $\uparrow$ & PSNR $\uparrow$
\\\midrule
uORF  & 0.4974 & 0.5347 & 0.4291 & 0.2417 & 0.6862 & 24.9712 \\ 
uORF + kmeans & \textbf{0.651} & \textbf{0.7346} & \textbf{0.5304} & \textbf{0.1894} & \textbf{0.7176} & \textbf{26.1833}\\
% uORF + kmeans (direct)\\
% uORF + kmeans (shared MLP)\\
uORF + PW & 0.5784 & 0.6943 & 0.4773 & 0.221 & 0.703 & 25.6277\\
% uORF + MS\\
\bottomrule
\end{tabular}}
\end{adjustbox}
\end{center}
\end{table}

\noindent\textbf{Results.} We show quantitative results in Table \ref{tab-novel-view-synthesis} and qualitative results in appendix C. Overall, our models outperform the original uORF consistently over all metrics. In particular, our models can better reconstruct the chairs pointed to the right direction while original uORF cannot build a clear shape for most chairs.
\section{Conclusion}
\label{conclusion}
We propose to learn an inductive slot initialization from the input instead of using a random initialization which is widely used in the prior works for the slot-based methods. To evaluate the importance of permutation symmetry over slots, we design various models with non-permutation symmetry, permutation invariance and permutation equivariance into consideration. In particular, our proposed permutation equivariant mean-shift model enables additional flexibility without requiring a fixed number of slots in advance, while it achieves notable improvements on the reconstructed perception details.

% \section*{Reproducibility Statement}
% We have provided the code repository for training and evaluation with the links to download the required datasets and pre-trained models. We have also provided the details of network architectures and implementations in Appendix. The researchers can use the provided repository to reproduce the results or evaluate on their own datasets. 

% \newpage
\bibliographystyle{unsrtnat}
\bibliography{egbib}

\newpage
\appendix
\setcounter{figure}{5}  
\setcounter{table}{3}  

In the appendix, we show the implementation details and qualitative results. 
% For more details, please visit our repository\footnote{Code, data and evaluation results are available at \url{https://github.com/bmvc2023id471/bmvc2023id471} for the review purpose.}.

\section{Implementation details}
\label{app_implementation}
\begin{figure}[htb!]
\centering
\includegraphics[width=0.9\linewidth]{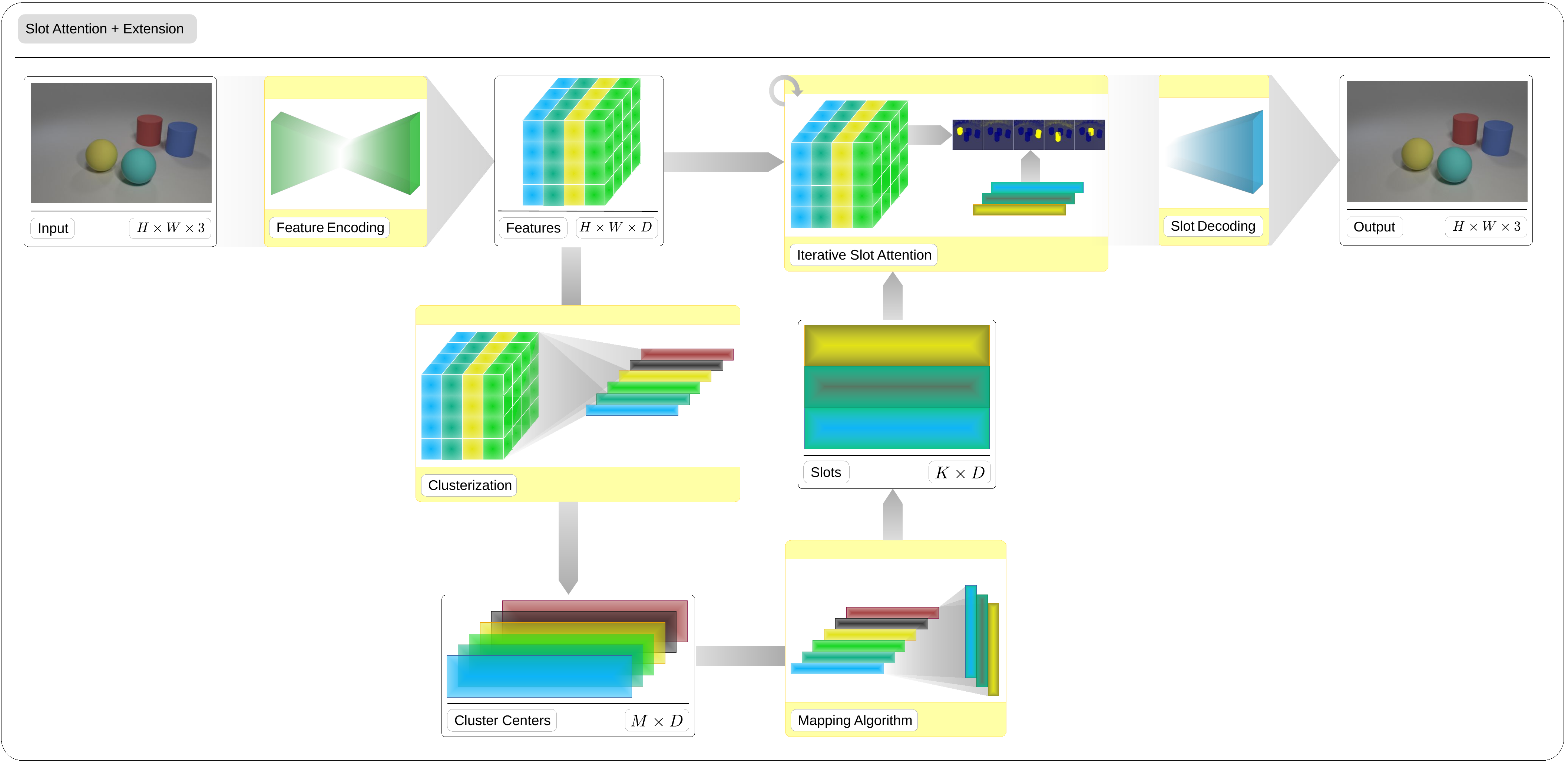}
\caption{The framework architecture for slot initialization for slto attention. The top row is the original architecture.}
\label{fig_app_overview_architecture}
\end{figure}

We show the implementation details and the architectures of different variants here.
\textbf{Slot Attention:} This section provides a detailed explanation of all the methods presented in chapter 2. The Slot Attention architecture in Figure \ref{fig_app_overview_architecture} is extended by a clusterization algorithm, that can be either k-means or mean shift, and by a mapping algorithm, being one of \textit{Direct}, \textit{Small MLP}, \textit{Large MLP} or \textit{Pseudoweights}. The encoder can be a U-Net or a size preserving convolution network. The extension initializes slots conditioned on the perceptual input and not like the original Slot Attention architecture from random gaussian distributions. During the iterative slot attention process, the initialized slots are updated to attend to certain feature pixels, while ignoring others. This is described by the bright yellow markings in the attention masks in Figure \ref{fig_app_overview_architecture}. The Slot attention uses three iterations to update the slots. Each slot is decoded into a rgb-image and an $\alpha$-mask. The renderer calculates, with a weighted sum, the output according to the slotwise rgb-image and the $\alpha$-mask.

\begin{figure}[htb!]
\centering
\includegraphics[width=0.9\linewidth]{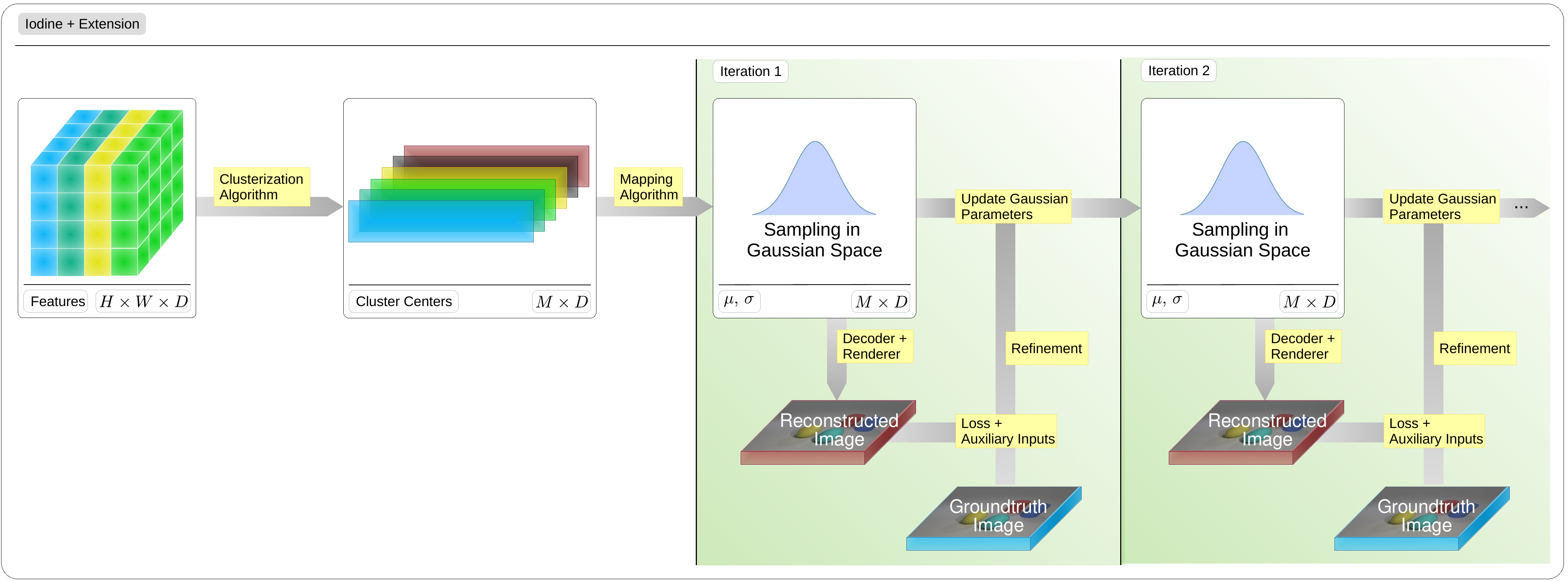}
\caption{The framework architecture for IODINE based extensions. The original starts directly at iteration 1 with slots drawn out of the standard gaussian distribution with $(\mu, \sigma) = (0, 1)$.}
\label{IODINE}
\end{figure}

\textbf{IODINE:} The extension for IODINE resemble the same structure as in the slot attention architecture in Figure \ref{IODINE}. The only difference is that the mapping algorithm has to map between the cluster centers of dimension $D$ to two parameters $\mu, \sigma$ of the Gaussian distribution. That is why \textit{Direct} mapping is impossible for IODINE. Slot initializations are now drawn out of the perceptual conditioned gaussian distribution and have dimension D. A decoder calculates, in the same fashion as for slot attention, for each slot a rgb-image and an $\alpha$-mask. The render outputs the reconstructed image, that will be compared to the groundtruth image to produce a loss. The loss is used in a refinement network, with auxiliary inputs, to update the gaussian parameters $\mu, \sigma$. This process is repeated five times.

\begin{figure}[htb!]
\centering
\includegraphics[width=0.9\linewidth]{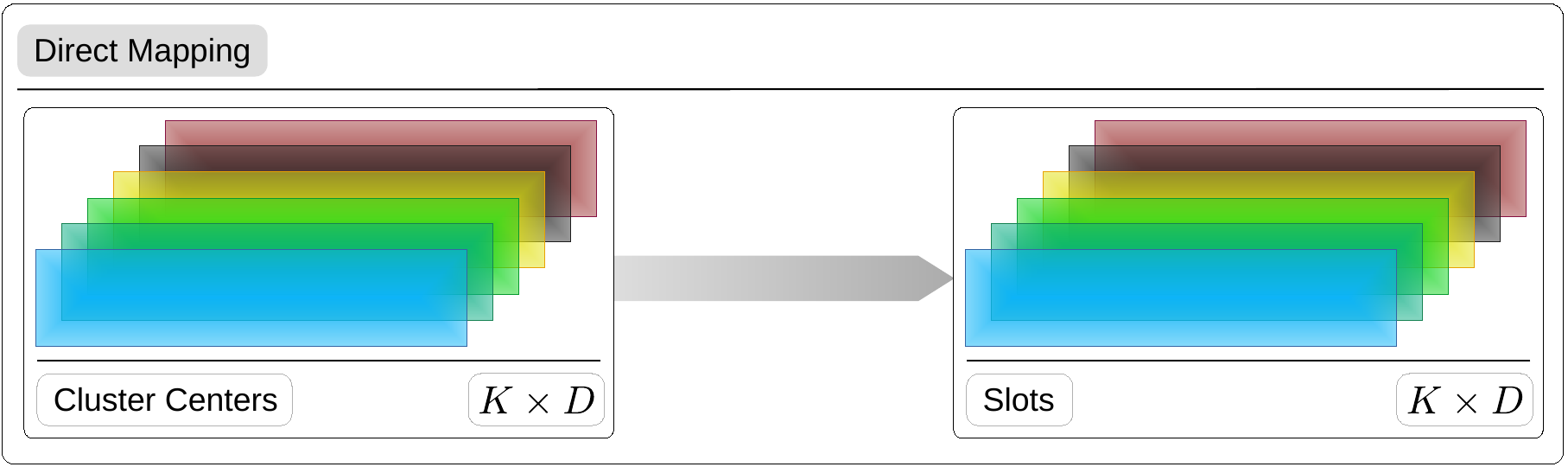}
\caption{The Direct mapping approach. Slots are identical to the cluster centers chosen by the clusterization algorithm.}
\label{fig_app_direct_architecture}
\end{figure}

\textbf{Direct mapping:} This simple permutation equivariant approach depicted in Figure \ref{fig_app_direct_architecture} directly injects the cluster centers determined by the clusterization algorithms into the slots. Since there is no mapping network involved, this approach can not be used for IODINE, because the cluster centers have to be mapped to two gaussian parameters $\mu, \sigma$.

\begin{figure}[htb!]
\centering
\includegraphics[width=0.9\linewidth]{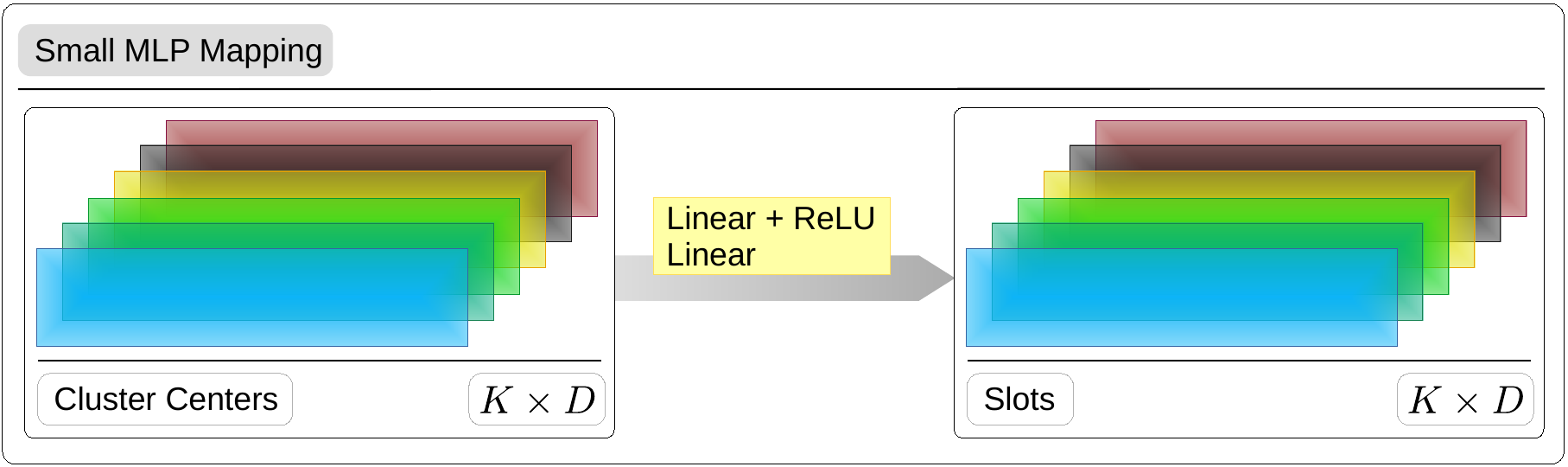}
\caption{The Small MLPs mapping approach. It extends the direct mapping approach by a nonlinear network between cluster centers and slots.}
\label{fig_app_smlp_architecture}
\end{figure}

\textbf{Small MLPs:} This mapping extends \textit{Direct}-mapping with a non linear network between the cluster centers and the slots, that is shared between all slots, as depicted in Figure \ref{fig_app_smlp_architecture}. The \textit{Direct}- and \textit{Small MLPs}-mapping are used for their simplicity and the permutation equivariance. But they can only translate between the same number of cluster centers and slots.

\begin{figure}[htb!]
\centering
\includegraphics[width=0.9\linewidth]{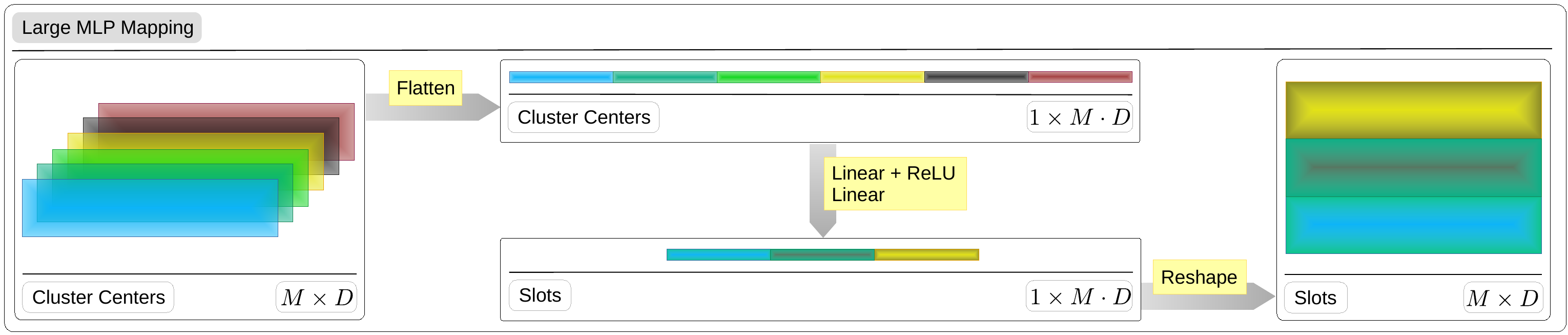}
\caption{The Large MLPs approach.}
\label{fig_app_lmlp_architecture}
\end{figure}

\textbf{Large MLPs:} This network maps between a different number of cluster centers and slots, as provided in Figure \ref{fig_app_lmlp_architecture}. The reason for this is to increase the sampling amount of cluster centers from the perceptual input without increasing the model size noticeably, which scales linear with the amount of slots. It is not shared between the slots and thus it is not permutation symmetric. The cluster centers are flattened into one large vector and then mapped to a flattened representation of the slots. These slots are then reshaped to $M \times D$. A drawback of this design is, that it can not generalize to more slots, like all other mapping networks, because of the fixed large MLPs.

\begin{figure}[htb!]
\centering
\includegraphics[width=0.9\linewidth]{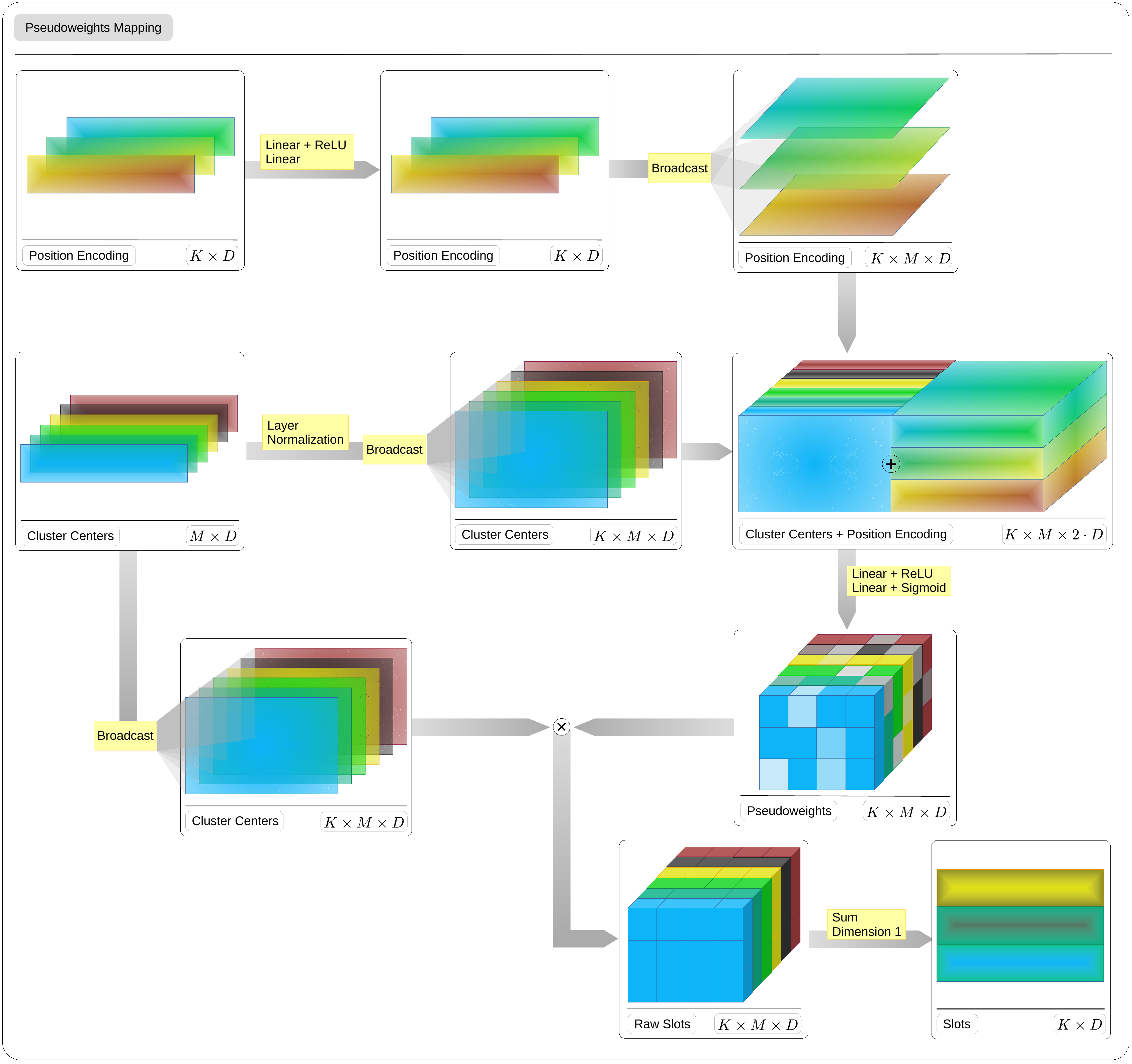}
\caption{The permutation invariant Pseudoweights mapping.}
\label{fig_app_pw_architecture}
\end{figure}

\begin{figure}[htb!]
\centering
\includegraphics[width=0.9\linewidth]{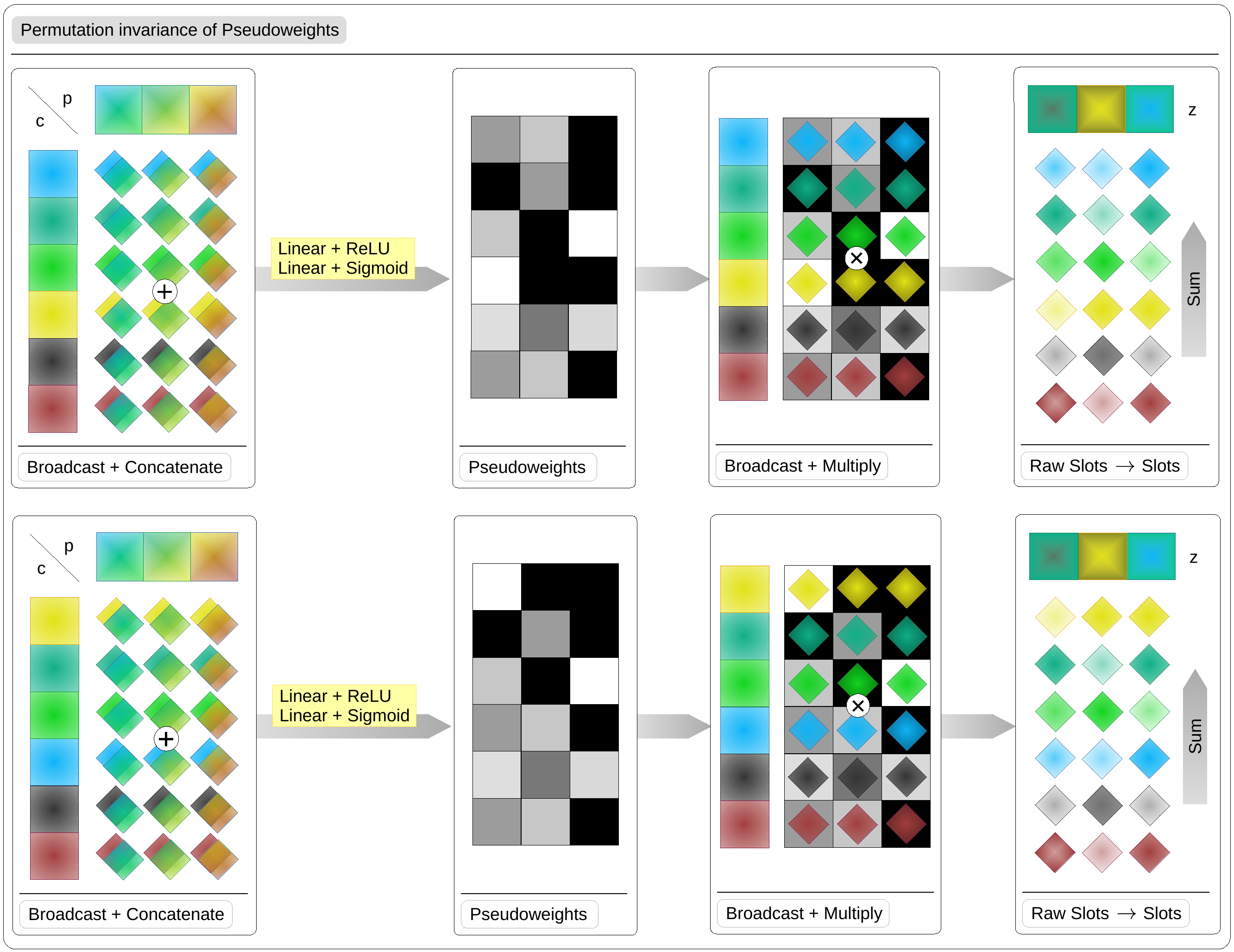}
\caption{The permutation invariant mapping between 6 cluster centers and 3 slots. For this example all slots and cluster centers are of dimension D=1, to keep it simple. The pseudoweights tensor has high values in black squares and low values in white squares. If the blue and yellow slot change their position, the slots won't change their initialization.}
\label{PermInvPsw}
\end{figure}

\textbf{Pseudoweights:} This algorithm incorporates several concepts into one mapping approach. It can map between a different amount of cluster centers and slots, while being able to generalize to more slots and keeping permutation invariance. It has to be permutation invariant, because it is ambiguous to define permutation equivariance between two not equally large sets. This mapping sorts cluster
centers into slots. It is aware in which slot it is, because of the position encoding of the K slots.
Thus the segregation network before the pseudoweights tensor can decide, if a cluster center should be sorted into a particular slot, then the weights in the
pseudoweights tensor will be high, other wise the weights will be low. This segregation network does the decision conditioned only on one cluster center and one position code for all possible $M \times K$ pairs. The last step calculates the weighted sum with the pseudoweights tensor and returns the initialized slots. An explanation of this process and a visual proof of permutation invariance is provided in Figure \ref{PermInvPsw}.

\textbf{Clusterization Algorithms:} The k-means algorithm used in the presented methods uses the k-means++ initialization, where the first center is randomly chosen and all other centers are initialized iteratively at the data point being the farthest away from all current initialized centers. If k-means is used with the \textit{Large MLP}, it requires a cluster dying prevention, because sometimes a cluster center will vanish, if all data points are closer to other cluster centers. In that case, a new cluster center is initialized with the k-means++ initialization. A pseudo code is provided in \ref{algo2}. The amount of cluster centers used in k-means is always initialized with the double amount of the maximum objects count in the dataset. So for CLEVR6, where there are up to six foreground objects and one background object, we initialize always 14 cluster centers at the start of k-means.

The mean shift algorithm is initialized with 20 cluster centers for all datasets, because after mean shift converges an algorithm called \textit{connected-components} is used to merge clusters centers, that are very close to each other in to one vector. This ability lets mean shift to determine the amount of slotsflexible. The hyper parameter $\epsilon$ is used to determine the radius in the \textit{connected-components}, where all cluster centers within the $\epsilon$-sphere are merged to one vector. Another hyper parameter used in mean-shift is $\sigma$ and is used to determine the bandwidth of the gaussian kernel. A detailed pseudo code is provided in \ref{algo1}. We determine the hyperparameters dependent on the weight initialization of the network, so that from the beginning of the training, the output amount of slots fluctuates between 1 and 20, but will never be always 20 or always 1. This happens if $\sigma$ or $\epsilon$ are too small, then mean shift will converge into every little mode, or if the hyperparameters are too large, then all cluster centers can merge into the same spot. 

\algnewcommand\algorithmicforeach{\textbf{for each}}
\algdef{S}[FOR]{ForEach}[1]{\algorithmicforeach\ #1\ \algorithmicdo}

\begin{algorithm}[h]
    \label{algo1}
    \centering
    \caption{K-means algorithm with cluster dying prevention, that reinitializes a new cluster center as soon as one vanishes.}
    \label{algorithm}
    \begin{algorithmic}[1]
        \State {$c_{i} \gets $ k-means++ initialization;\, $i \leq N$}
        \Repeat
        \ForEach{$c_{i}$}
        \State $C_{i} = \{x_{j}: d(x_{j}, c_{i}) \leq d(x_{j}, c_{k}); \forall x_{j} \land \forall k \neq i\}$
        \EndFor
        \ForEach{$C_{i}$}
        \If{$C_{i} = \varnothing$}
        \State {$c_{inew} \gets$ k-means++ reinitialization}
        \Else
        \State {$c_{inew} = \sum_{c_{i} \in C_{i}} \frac{c_{i}}{|C_{i}|}$}
        \EndIf
        \EndFor
        \If {$d(c_{i}, c_{inew}) \leq tolerance \, \forall i$}
        \State \textbf{Return}  {$c_{inew}$}
        \EndIf
        \Until {max iterations}
        \State \textbf{Return}  {$c_{inew}$}
    \end{algorithmic}
\end{algorithm}

\begin{algorithm}[H]
    \label{algo2}
    \centering
    \caption{Mean shift algorithm, with the hyperparameters $\epsilon$ used in the connected-components algorithm and $\sigma$ used in the gaussian kernel function.}\label{algorithm}
    \begin{algorithmic}[1]
        \For{$n \in {1, ..., N}$}
        \State{$x \gets x_{n}$}
        \Repeat
        \State {$\forall n: p(n|x) \gets \frac{exp(-0.5 ||\frac{x-x_{n}}{\sigma}||^2)}{\sum_{n'=1}^{N} exp(-0.5 ||\frac{x-x_{n}}{\sigma}||^2})$}
        \State {$x \gets \sum_{n'=1}^{N} p(n|x) \cdot x_{n}$}
        \Until{stop}
        \State{$z_{n} \gets x$}
        \EndFor
        \State{connected-components($\{z_{n}\}_{n=1}^{N}, \epsilon$)}
    \end{algorithmic}
\end{algorithm}

\section{Visualizations on object discovery task}
\label{app_sec_visualization}
We show some qualitative evaluation examples for the object discovery task in Fig.13-20. 

\begin{figure}[htb!]
\centering
\includegraphics[width=0.9\linewidth]{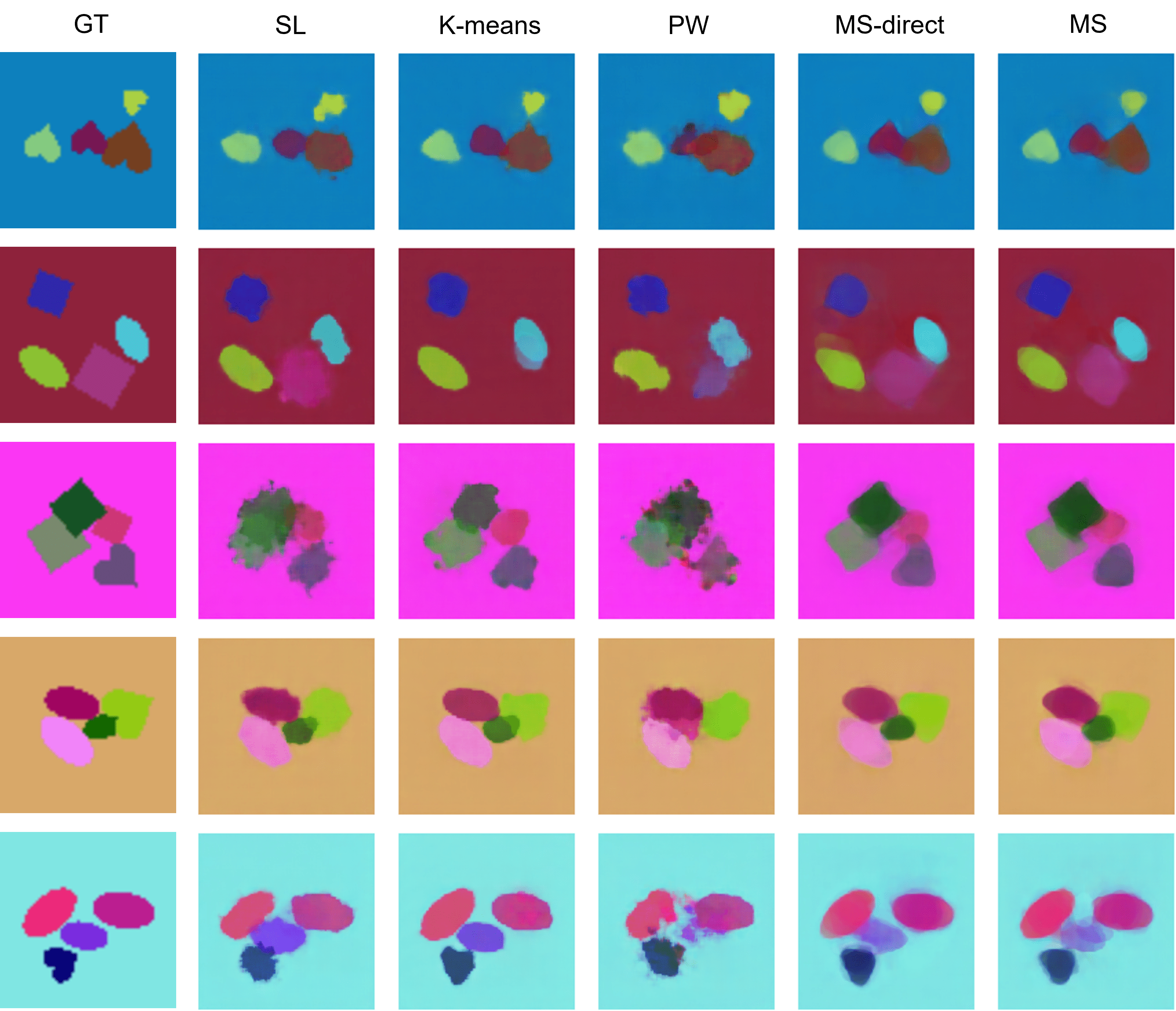}
\caption{Qualitative results on MDS dataset.}
\label{fig_app_sl_mds}
\end{figure}

\begin{figure}[htb!]
\centering
\includegraphics[width=0.9\linewidth]{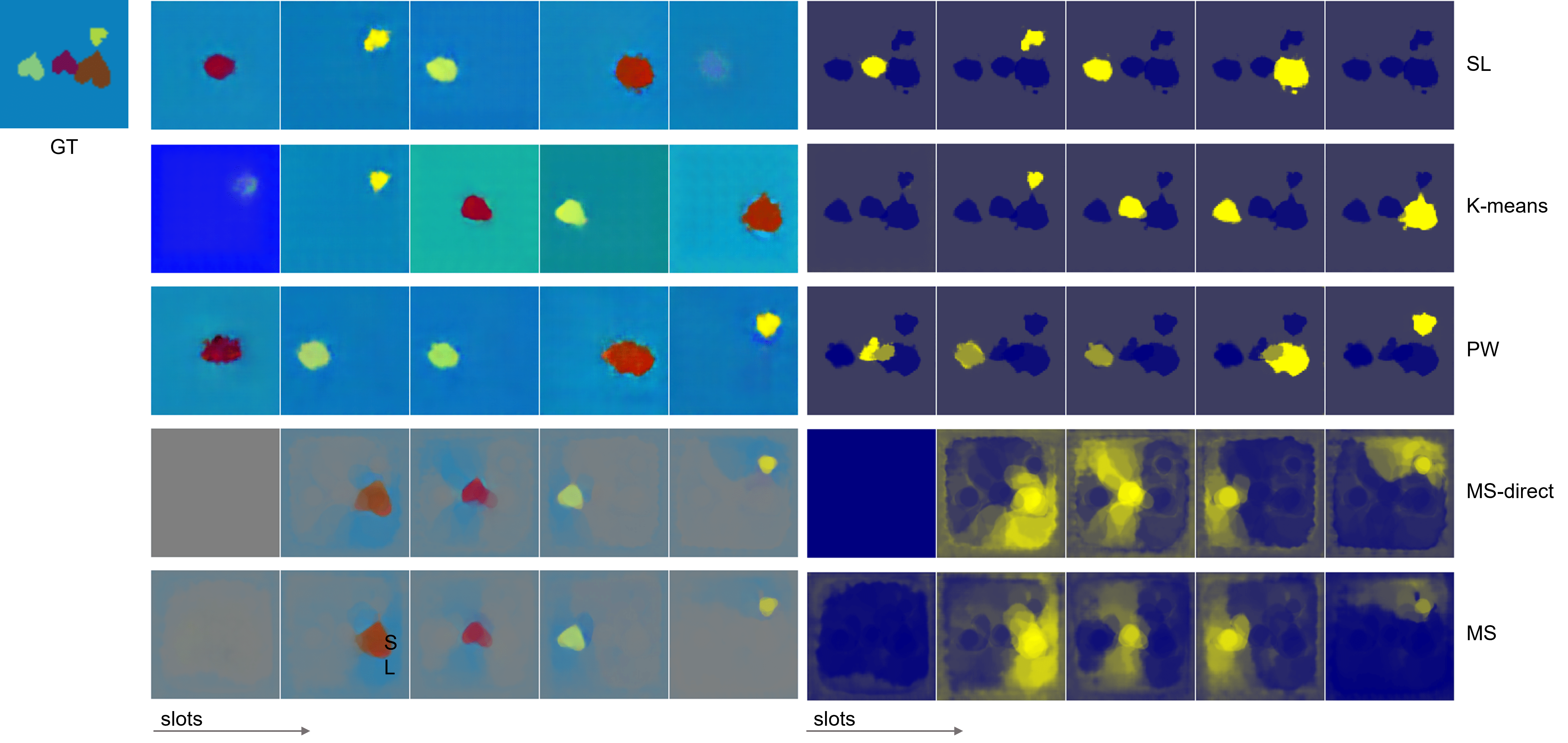}
\caption{The slot-wise predicted masks and reconstructed scenes on MDS dataset.}
\label{fig_app_sl_mds_masks}
\end{figure}

\begin{figure}[htb!]
\centering
\includegraphics[width=0.9\linewidth]{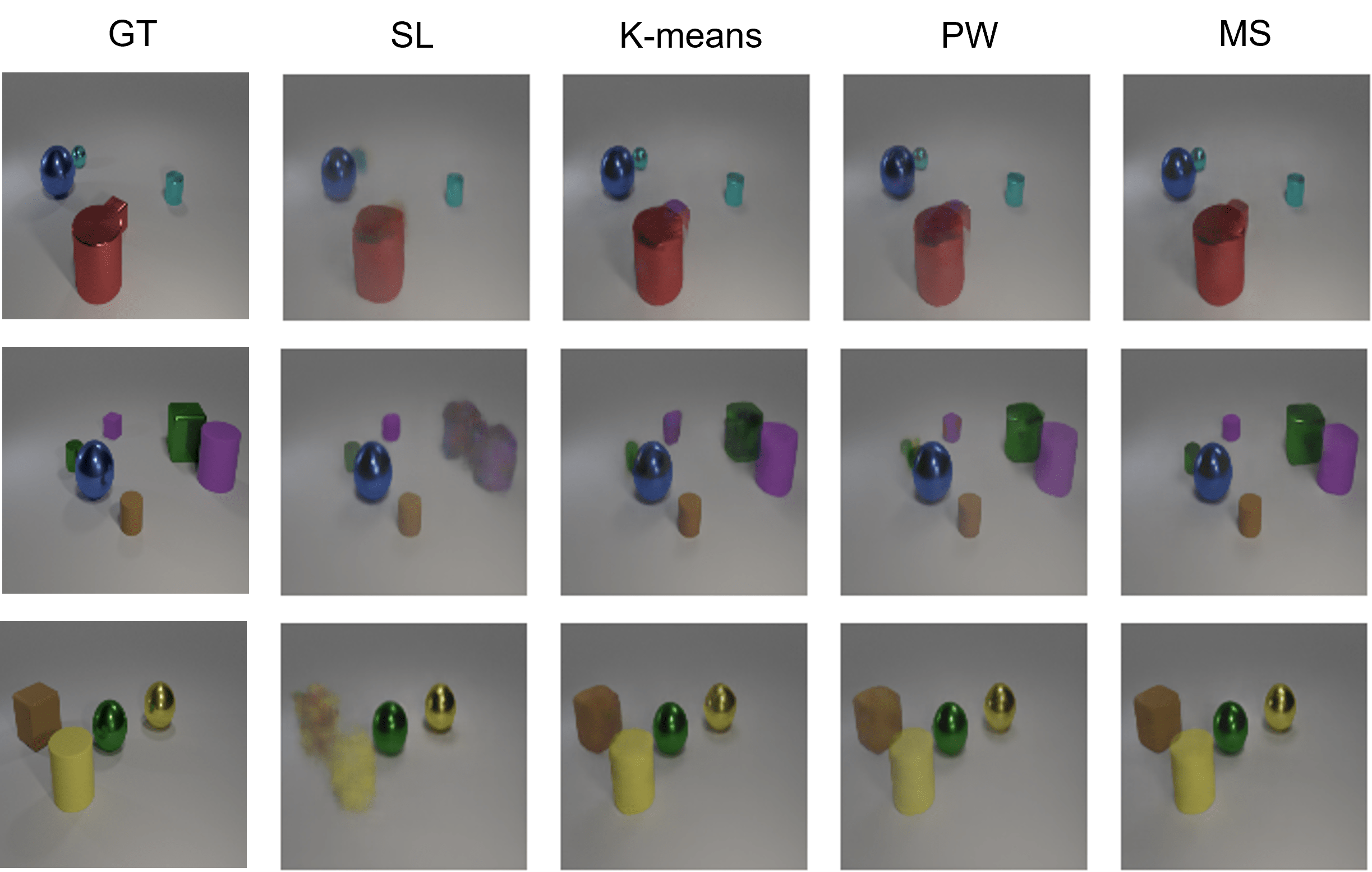}
\caption{The original Slot Attention model struggles with overlapped objects.}
\label{fig_app_sl_clevr6}
\end{figure}

\begin{figure}[htb!]
\centering
\includegraphics[width=0.9\linewidth]{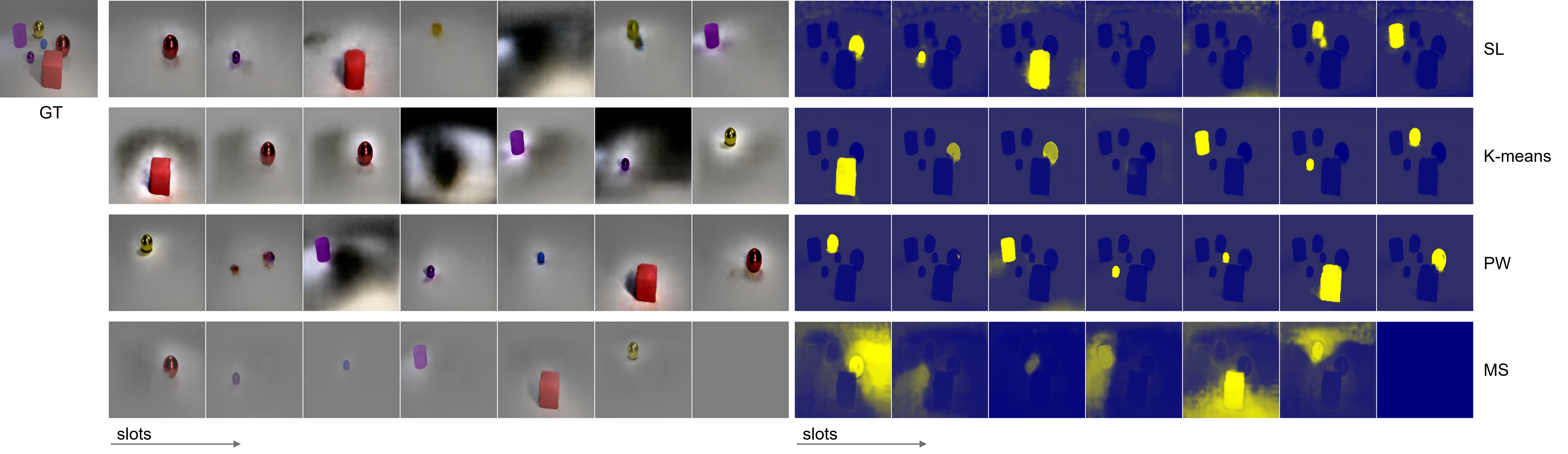}
\caption{The slot-wise predicted masks and reconstructed scenes on CLEVR6 dataset.}
\label{fig_app_sl_clevr6_masks}
\end{figure}

\begin{figure}[htb!]
\centering
\includegraphics[width=0.9\linewidth]{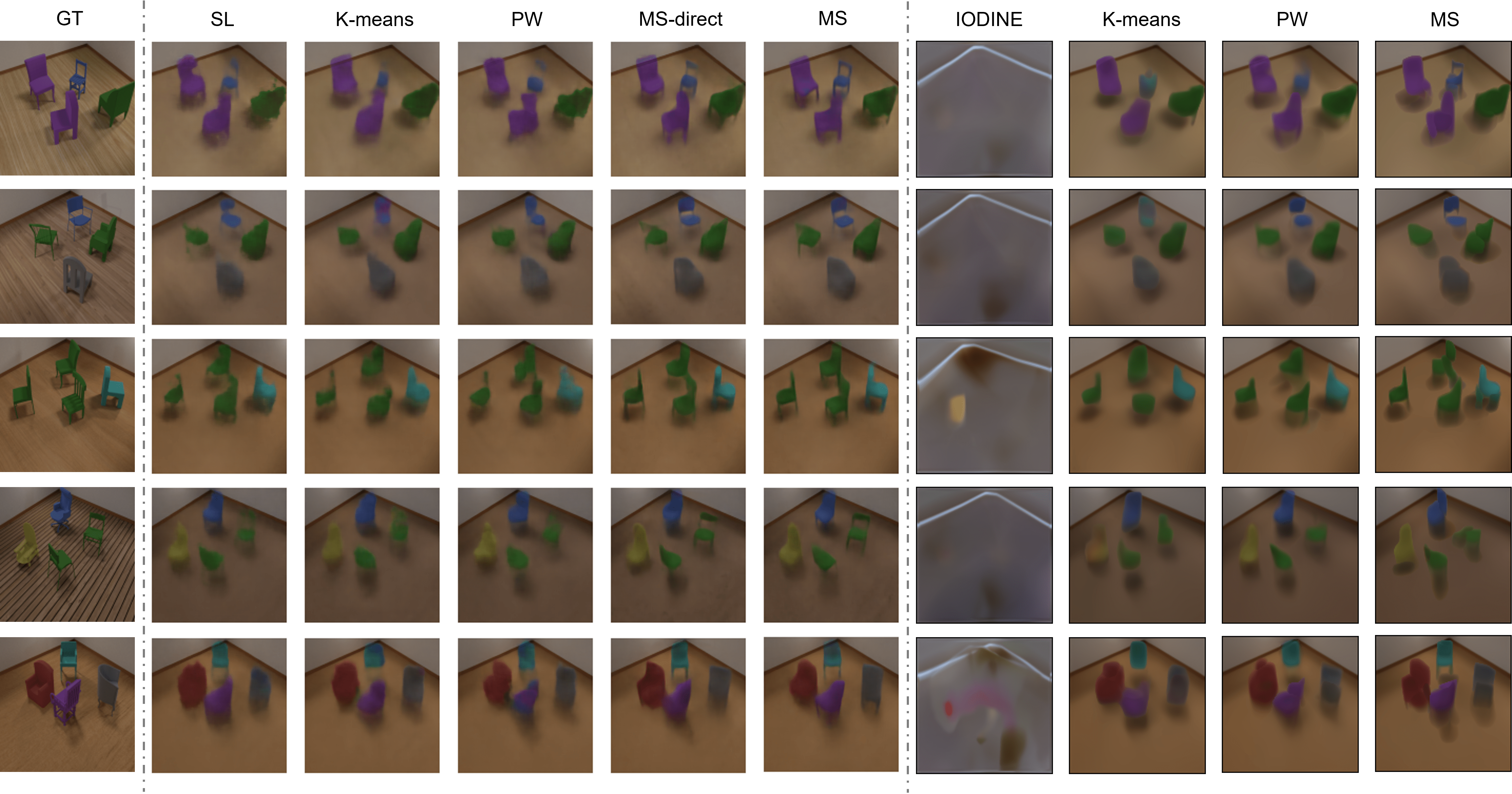}
\caption{Qualitative results on Chairs dataset.}
\label{fig_app_sl_chairs}
\end{figure}

\begin{figure}[htb!]
\centering
\includegraphics[width=0.9\linewidth]{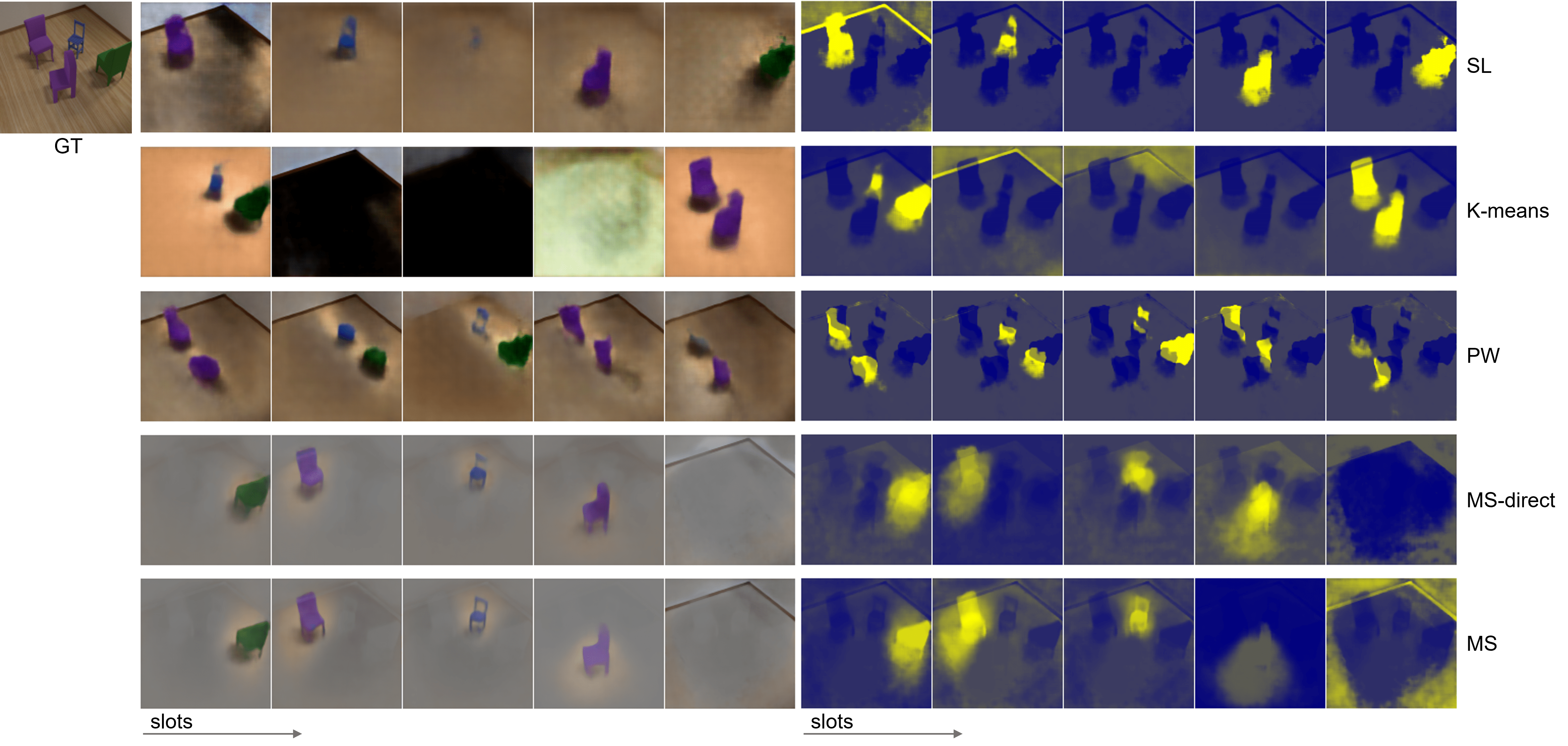}
\caption{The slot-wise predicted masks and reconstructed scenes on Chairs dataset.}
\label{fig_app_sl_chairs_masks}
\end{figure}

\begin{figure}[htb!]
\centering
\includegraphics[width=0.9\linewidth]{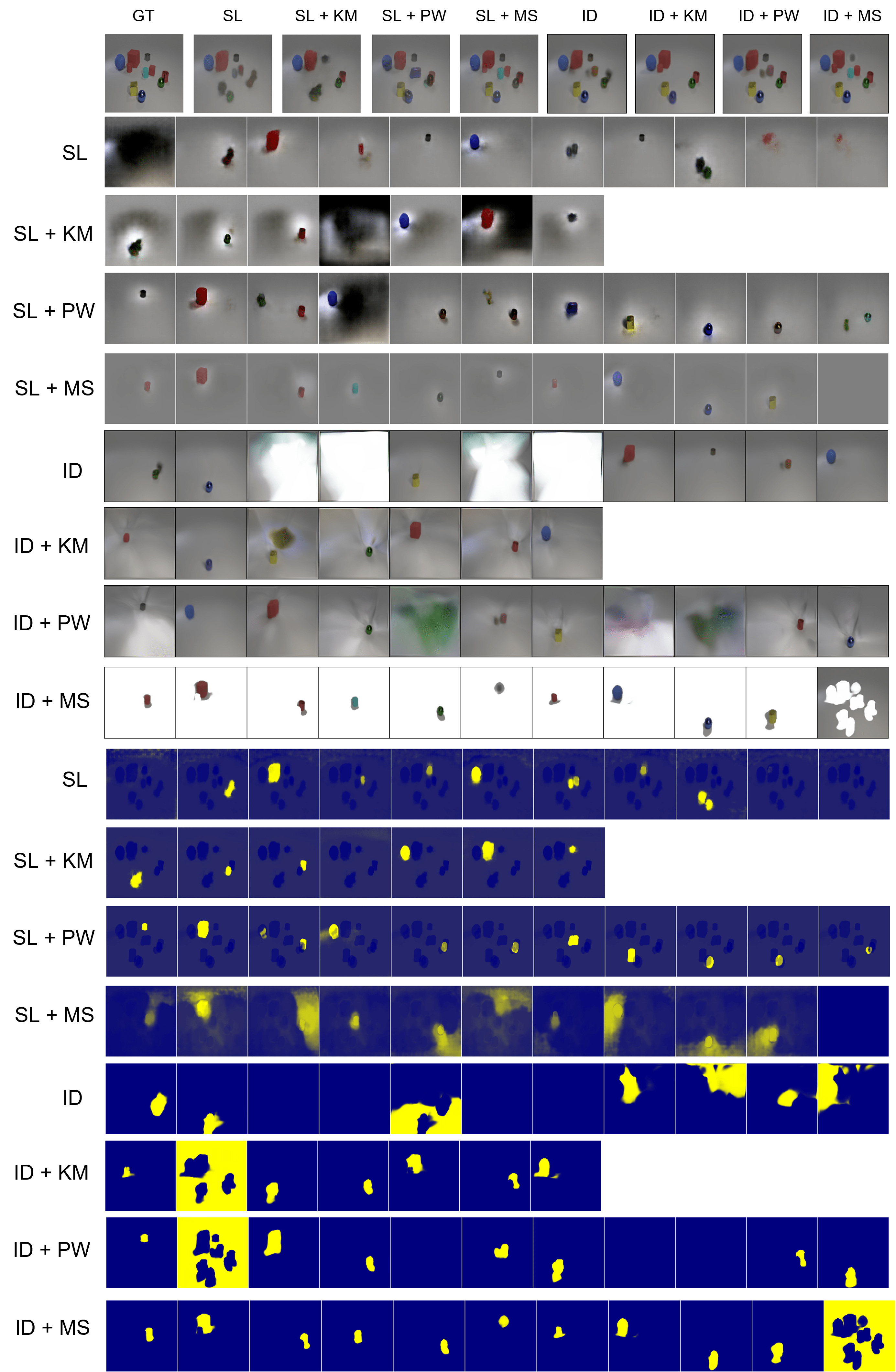}
\caption{Qualitative comparison of generalization on CLEVR10 while the models are trained with CLEVR6.}
\label{fig_app_sl_generalization_clvr10_01}
\end{figure}

\begin{figure}[htb!]
\centering
\includegraphics[width=0.9\linewidth]{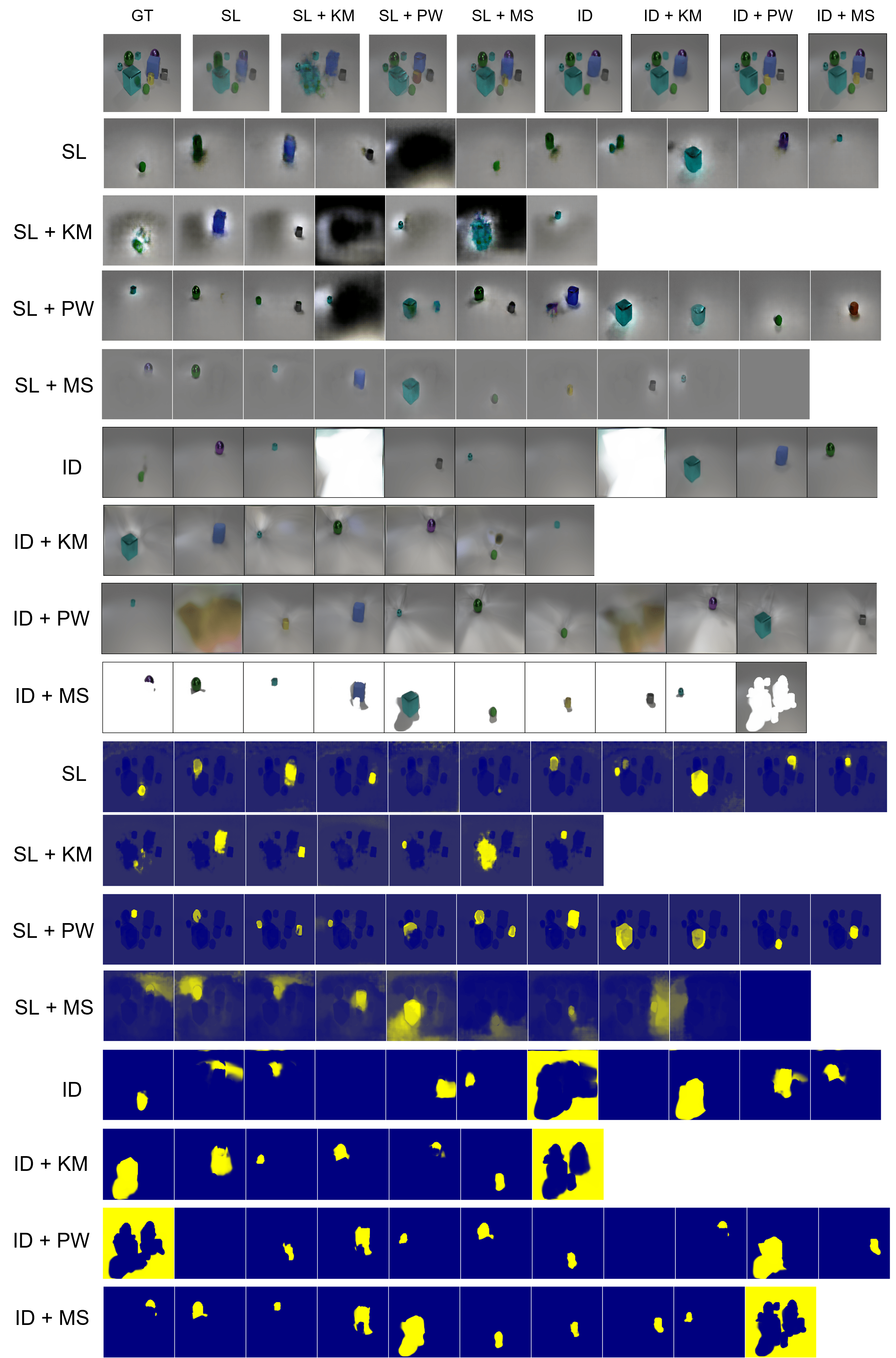}
\caption{Another qualitative comparison of generalization on CLEVR10.}
\label{fig_app_sl_generalization_clvr10_02}
\end{figure}

\section{Visualizations on novel view synthesis task}
\label{app_sec_visualization_novel_view_synthesis}
\begin{figure}[hbt!]
\centering
\includegraphics[width=0.9\linewidth]{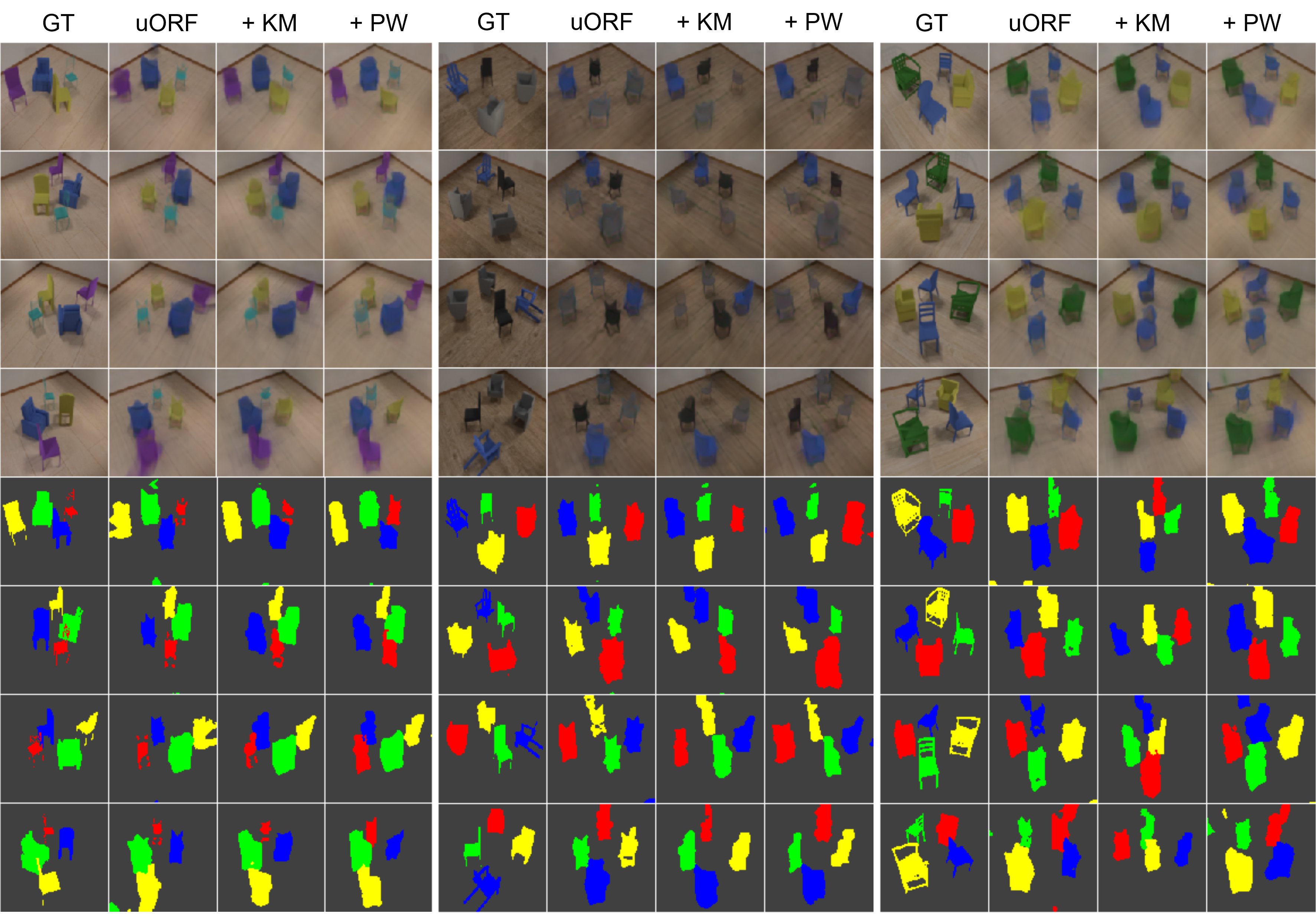}
\caption{Qualitative results on novel view synthesis. Our models can represent the chairs with more details than the original uORF.}
\label{fig_uorf_results}
\end{figure}

We visualize the examples of novel view synthesis tasks in Figure \ref{fig_uorf_results}.

\end{document}